\documentclass[Afour,sageh,times]{sagej}

\usepackage{moreverb,url}

\usepackage[colorlinks,bookmarksopen,bookmarksnumbered,citecolor=blue,urlcolor=blue, linkcolor=blue]{hyperref}

\usepackage{graphics} %
\usepackage{graphicx}
\usepackage{bbding}
\usepackage{amsmath,nccmath} %
\usepackage{amssymb}  %
\usepackage{acronym}
\usepackage{mathrsfs}
\usepackage{nicefrac}
\usepackage{subfigure}
\usepackage{pifont}  %
\usepackage{afterpage}  %
\usepackage[T1]{fontenc}
\usepackage{tgbonum}
\usepackage{booktabs}    %
\usepackage{multirow}    %
\usepackage{array}       %

\usepackage{wasysym}     %

\newcommand\BibTeX{{\rmfamily B\kern-.05em \textsc{i\kern-.025em b}\kern-.08em
T\kern-.1667em\lower.7ex\hbox{E}\kern-.125emX}}

\def\volumeyear{2026}

\begin{document}

\runninghead{Malone et al.}

\title{FRED: A Multi-Modal Autonomous Driving Dataset for Flooded Road Environments}

\author{Connor Malone\affilnum{1}, Sébastien Demmel\affilnum{1} and Sébastien Glaser\affilnum{1}}

\affiliation{\affilnum{1}ARC Training Centre for Automated Vehicles in Rural and Remote Regions (AVR3), Queensland University of Technology}

\corrauth{Connor Malone,
Queensland University of Technology, Australia}

\email{cj.malone@qut.edu.au}

\begin{abstract}
The Flooded Road Environments Dataset (FRED) is, to our knowledge, the first multi-modal autonomous driving dataset specifically targeting the collection of data from scenarios involving water hazards on the road. The dataset contains images from a 2.3 MP FLIR Blackfly USB3 camera, 64-beam 360$^\circ$ point clouds from an Ouster OS1-64 LiDAR, and data from an iXblue ATLANS-C IMU corrected by a Geoflex RTK GNSS, from five separate locations captured both during and after flooding events. The data has been released in two formats: a KITTI-style format for easy integration with existing data tools, and the RTMaps format for direct replay of the vehicle's data capture. We provide semantic labels to enable the training and evaluation of both single-sensor and sensor-fusion methods for water hazard detection. Position and velocity, as well as data captured under dry conditions, are provided to enable the development of location-based detection methods that may incorporate maps, and to evaluate other tasks such as localisation and SLAM.
\end{abstract}


\maketitle

\renewcommand{\thefootnote}{\arabic{footnote}}

\section{Introduction}
\label{sec:introduction}
\noindent Autonomous vehicles are being increasingly adopted and deployed in industrial, freight, and ride-hailing applications that operate within populated environments (\cite{di2024autonomous, jones2025market}). Accordingly, recent research has focused more on developing perception and localisation systems that remain robust and safe in adverse conditions such as rain and snow, and in night-time scenarios (\cite{zhang2023perception, Malone_2025_ICCV, nahata2023exploring, almalioglu2022deep, malone2022improving, bruggemann2023refign}). A critical component for enabling this research is the collection, labelling, and release of datasets capturing the operation of autonomous vehicles in these conditions and scenarios.

Water hazards are a challenging perception problem for robots and autonomous vehicles (\cite{wijayathunga2023challenges}). Both image-based and LiDAR-based approaches struggle to consistently and robustly detect bodies of water in road environments due to the large variation in appearance and the complex interaction between LiDAR beams and water (\cite{matthies2003detecting, rankin2011daytime, goodin2019predicting, zang2019impact}). Failing to detect water hazards, such as flooded roads, can have catastrophic consequences for both the vehicle and any cargo or passenger being transported. However, despite the safety-critical nature of this task, there are limited datasets containing labelled data from vehicles encountering water hazards on the road, and even fewer specifically targeting these scenarios. Consequently, there is also limited research developing perception systems that can robustly detect them.

\begin{figure}[t]
    \centering
    \includegraphics[width=\linewidth]{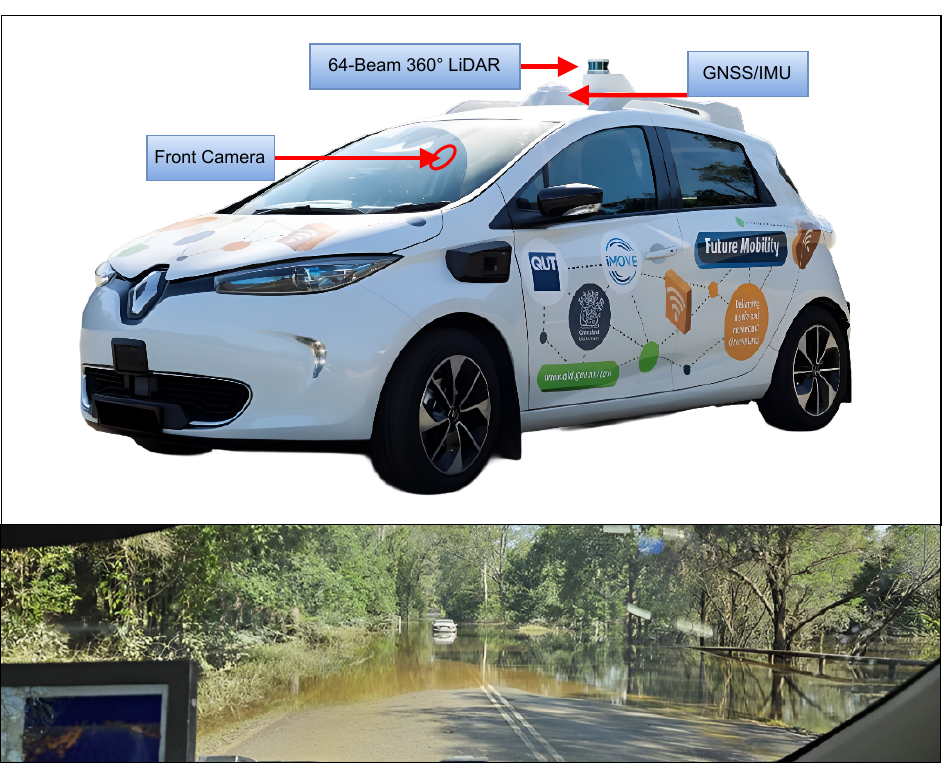}
    \caption{\textbf{Above:} Our \textit{Zoe 2} data collection vehicle, including front and rear FLIR Blackfly cameras, an Ouster OS1-64 LiDAR, and an iXblue ATLANS-C IMU corrected by a Geoflex RTK GPS. \textbf{Below:} A sample of the dataset demonstrating the danger of undetected water hazards.}
    \label{fig:zoe2}
    \vspace{-\baselineskip}
\end{figure}

In this work, we present the Flooded Road Environments Dataset (FRED) to encourage and enable the advancement of perception systems for detecting water hazards, such as flooded roads. The dataset was collected by capturing the output of an autonomous vehicle sensor stack at five separate locations, both during and after flooding events. It contains $\approx$ 5340 data samples, which each include 2.3 MP images from both front and rear facing cameras, 360$^\circ$ 64-beam point-clouds, and GNSS corrected IMU data, all synchronised using centralised timestamps. We provide semantic labels for image and (using projection) point cloud samples to identify water hazards in each sequence.

The main objective of this dataset is to support the development of various types of methods for detecting water hazards, including image-based, LiDAR-based, map-based, and sensor fusion approaches. In addition, we provide position information to allow the dataset to also be used for evaluating localisation and SLAM systems in these scenarios. Concretely, our contributions are:

\begin{itemize}
    \item A multi-modal dataset capturing five separate locations, both during and after flooding events.
    \item Semantic labels for developing robust camera, LiDAR, and sensor fusion-based water hazard detection methods.
    \item GNSS corrected IMU data for assisting in the generation of maps and map-based water hazard detection approaches, as well as for evaluating localisation methods.
    \item Two formats of the dataset, including a KITTI-style format and the native RTMaps data capture format.
    \item Benchmark results for recent image-based approaches in semantic segmentation and localisation tasks.
\end{itemize}

The dataset is available here\footnote{\scriptsize{\url{https://huggingface.co/datasets/CMalone-Jupiter/FRED}}}\textsuperscript{,}\footnote{\scriptsize{\url{https://data.researchdatafinder.qut.edu.au/dataset/flooded-road-environments}}}, and a purpose-developed software development kit has been made publicly available already\footnote{\scriptsize{\url{https://github.com/AVR3-Training-Centre/python-FRED}}}\textsuperscript{,}\footnote{\scriptsize{\url{https://huggingface.co/spaces/CMalone-Jupiter/python-FRED}}}.

The manuscript will proceed as follows, related works on autonomous vehicle datasets in Section~\ref{sec:related-works}, information on the data collection and sensors in Sections~\ref{sec:data-collection} and \ref{sec:sensors}, the dataset formats are documented in Section~\ref{sec:dataset-format}, calibration, annotations, and the development kit are discussed in Sections~\ref{sec:calibration}--\ref{sec:development-kit}, and finally benchmark metrics and the conclusions are given in Sections~\ref{sec:benchmark-metrics} and \ref{sec:conclusion}.
\section{Related Works}
\label{sec:related-works}
\subsection{Autonomous Vehicle Datasets}
Datasets are a crucial resource for the development and advancement of autonomous vehicle systems. In particular, research communities investigating perception and localization tasks require substantial amounts of data to reliably train, validate, and evaluate modern deep learning approaches. This has resulted in the creation of many widely utilised benchmark datasets, such as the KITTI~(\cite{geiger2013vision}) and Cityscapes~(\cite{Cordts_2016_CVPR}) datasets for perception, and the Oxford RobotCar~(\cite{maddern20171}) and NuScenes~(\cite{caesar2020nuscenes}) datasets for SLAM/localization.

\begin{table*}
  \centering
  \caption{Common autonomous vehicle datasets used in perception and localization research. The \CIRCLE\ symbol indicates a dataset satisfies the criteria in the corresponding column, and the \Circle\ symbol indicates it does not. The \RIGHTcircle\ symbol is used in the case of semantic point cloud labels to indicate where semantic image labels could be transferred to point clouds using point projection.}
      \renewcommand\arraystretch{0.89}
      \renewcommand\tabcolsep{3pt}
      \scriptsize
      \begin{tabular}{clcccccccc}
        \toprule[0.03cm]
        \multirow{2}{*}{\rotatebox{90}{\textbf{}}} 
        & \multirow{2}{*}{\textbf{Dataset}} 
        & \multicolumn{4}{c}{\textbf{Sensors Modality}} 
        & \multicolumn{2}{c}{\textbf{Semantic Labels}} 
        & \textbf{Adverse Weather}
        & \multirow{2}{*}{\textbf{Water Hazards}} \\
        \cmidrule(lr){3-6}
        \cmidrule(lr){7-8}
        & & IMU 
          & Camera
          & LiDAR
          & GPS & Image & Point Cloud & \textbf{(Night/Rain/Flood/Snow)} \\
        \midrule[0.03cm]

        \multicolumn{10}{c}{General Autonomous Driving Datasets} \\

        & NuScenes~(\cite{caesar2020nuscenes})
        & \CIRCLE & \CIRCLE & \CIRCLE & \CIRCLE & \Circle & \CIRCLE & \CIRCLE & \Circle \\

        & Argoverse~(\cite{chang2019argoverse})
        & \CIRCLE & \CIRCLE & \CIRCLE & \CIRCLE & \Circle & \CIRCLE & \CIRCLE & \Circle \\

        & Waymo Open Perception~(\cite{sun2020scalability})
        & \CIRCLE & \CIRCLE & \CIRCLE & \CIRCLE & \CIRCLE & \CIRCLE & \Circle & \Circle \\

        & Ford Multi-AV~(\cite{agarwal2020ford})
        & \CIRCLE & \CIRCLE & \CIRCLE & \CIRCLE & \Circle & \Circle & \CIRCLE & \Circle \\

        \midrule[0.03cm]

        \multicolumn{10}{c}{SLAM/Localization Focused Datasets} \\
        
        & Oxford RobotCar~(\cite{maddern20171})
        & \CIRCLE & \CIRCLE & \CIRCLE & \CIRCLE & \Circle & \Circle & \CIRCLE & \Circle \\

        & Nordland~(\cite{neubert2015superpixel})
        & \Circle & \CIRCLE & \Circle & \Circle & \Circle & \Circle & \CIRCLE & \Circle \\

        & KAIST-Complex Urban~(\cite{jeong2019complex})
        & \CIRCLE & \CIRCLE & \CIRCLE & \CIRCLE & \Circle & \Circle & \Circle & \Circle \\

        & MIT DARPA~(\cite{huang2010high})
        & \CIRCLE & \CIRCLE & \CIRCLE & \CIRCLE & \Circle & \Circle & \Circle & \Circle \\

        \midrule[0.03cm]

        \multicolumn{10}{c}{Perception Focused Datasets} \\

        & Cityscapes~(\cite{Cordts_2016_CVPR})
        & \Circle & \CIRCLE & \Circle & \Circle & \CIRCLE & \Circle & \Circle & \Circle \\

        & KITTI~(\cite{geiger2013vision})
        & \CIRCLE & \CIRCLE & \CIRCLE & \CIRCLE & \CIRCLE & \CIRCLE & \Circle & \Circle \\

        & Apolloscape~(\cite{huang2018apolloscape})
        & \CIRCLE & \CIRCLE & \CIRCLE & \CIRCLE & \CIRCLE & \CIRCLE & \Circle & \Circle \\

        \midrule[0.03cm]

        \multicolumn{10}{c}{Adverse Condition focused Datasets} \\

        & Boreas~(\cite{burnett2023boreas})
        & \CIRCLE & \CIRCLE & \CIRCLE & \CIRCLE & \Circle & \CIRCLE & \CIRCLE & \Circle \\
    
        & SFU Mountain~(\cite{bruce2015sfu})
        & \CIRCLE & \CIRCLE & \CIRCLE & \CIRCLE & \Circle & \Circle & \CIRCLE & \Circle \\

        & CADC~(\cite{pitropov2021canadian})
        & \CIRCLE & \CIRCLE & \CIRCLE & \CIRCLE & \Circle & \CIRCLE & \CIRCLE & \Circle \\

        & Dark Zurich~(\cite{sakaridis2019guided})
        & \Circle & \CIRCLE & \Circle & \Circle & \CIRCLE & \Circle & \CIRCLE & \Circle \\

        \midrule[0.03cm]

        \multicolumn{10}{c}{Water Hazard Focused Datasets} \\

        \midrule[0.03cm]

        & Puddles-1000~(\cite{han2018single})
        & \Circle & \CIRCLE & \Circle & \Circle & \CIRCLE & \Circle & \Circle & \CIRCLE \\
    
        & Night-Puddle~(\cite{zhang2024agsenet})
        & \Circle & \CIRCLE & \Circle & \Circle & \CIRCLE & \Circle & \Circle & \CIRCLE \\
    
        & Semantic Spray++~(\cite{piroli2024semanticspray++})
        & \Circle & \CIRCLE & \CIRCLE & \Circle & \CIRCLE & \CIRCLE & \CIRCLE & \Circle \\
    
        & Kaggle (Image) Datasets
        & \Circle & \CIRCLE & \Circle & \Circle & \CIRCLE & \Circle & \CIRCLE & \CIRCLE \\

        \midrule[0.03cm]
    
        & \textbf{Ours (FRED)}
        & \CIRCLE & \CIRCLE & \CIRCLE & \CIRCLE & \CIRCLE & \RIGHTcircle & \CIRCLE & \CIRCLE \\
    
        \bottomrule[0.03cm]
      \end{tabular}
    \label{tab:related_work_dataset_reduced}
    \vspace{-0.2cm}
\end{table*}

Each dataset typically provides data from a unique combination of sensors, which could include a camera, LiDAR, IMU, and/or GPS. Often, datasets such as Argoverse~(\cite{chang2019argoverse}), Ford Multi-AV~(\cite{agarwal2020ford}) or Kaist-Complex Urban~(\cite{jeong2019complex}), provide data from a full autonomy stack (all of the above sensors) to enable SLAM and localization research. Other datasets, such as Waymo Open Perception~(\cite{sun2020scalability}) or Apolloscape~(\cite{huang2018apolloscape}), additionally include semantic annotations of image and/or point cloud data to support the development of autonomous vehicle perception systems.

\subsection{Adverse Environmental Conditions}
Recently, state-of-the-art perception and localization systems have been able to achieve sufficiently high accuracy and robustness under \textit{normal}\footnote{Clear, well-illuminated conditions with up to moderate fluctuations in environmental conditions} operating conditions for companies to begin offering automated taxi services. Accordingly, it is increasingly important for researchers to address performance in adverse environmental conditions such as nighttime, snow, dust, heavy rain, and flooding to ensure safe operation. There is now several datasets available for perception and localization tasks that include data captured at night or in snowy, rainy, or even foggy conditions. Some, such as Boreas~(\cite{burnett2023boreas}), SFU Mountain~(\cite{bruce2015sfu}) and Canadian Adverse Driving Conditions (CADC)~(\cite{pitropov2021canadian}), provide recordings from complete autonomy stacks, whereas others, such as Nordland~(\cite{neubert2015superpixel}) and Dark Zurich~(\cite{sakaridis2019guided}), only provide a single sensor modality. However, despite the selection of datasets focused on adverse environmental conditions and several reported incidents of autonomous vehicles failing to safely navigate flood waters, there is a lack of datasets and research addressing flooded roads. Table~\ref{tab:related_work_dataset_reduced} summarises the characteristics of relevant datasets, including commonly used, adverse weather-focused, and water hazard-specific datasets.

\subsection{Water Hazards}
Detecting water hazards, such as flooded roads, using autonomous vehicles has been a challenging perception task for a long time. The Jet Propulsion Lab (JPL) at NASA published several works in the early 2000s investigating the detection of water hazards from unmanned ground vehicles (\cite{matthies2003detecting, rankin2006daytime, rankin2010daytime, rankin2010evaluating, rankin2011daytime, rankin2012water}). However, despite the significant body of research, no public dataset was released with these publications for future research to utilise and benchmark against.

More recently, \cite{han2018single} published a deep learning based approach to water hazard detection that combined Fully Convolutional Networks with an attention mechanism specifically designed to target areas in an image with reflections. To accompany the proposed approach, the authors released the Puddle-1000 dataset which contained 985 images with semantic annotations of roads covered and/or surrounded by pools of water. The Puddle-1000 dataset has become one of the main benchmark datasets used for research into the detection of water hazards. \cite{zhang2024agsenet} later contributed a similar dataset captured at nighttime, referred to as Night-Puddle. However, neither of these datasets includes LiDAR or position information, and both have become virtually inaccessible to researchers. The Puddle-1000 dataset unfortunately became inaccessible after the decommissioning of the CloudStor data storage service. Whereas, the Night-Puddle dataset appears to only be accessible through a Chinese-based service which is sometimes blocked in particular countries and requires a Chinese phone number to create an account.

The proposed FRED dataset addresses this critical lack of publicly available water hazard datasets, especially with data from a full autonomous vehicle sensor stack. It will encourage and enable more research into the challenging perception task of detecting flooded roads.

\setlength\tabcolsep{1.5pt}
\begin{figure*}
    \centering
    \begin{tabular}{cccc}
    \includegraphics[width=0.24\linewidth]{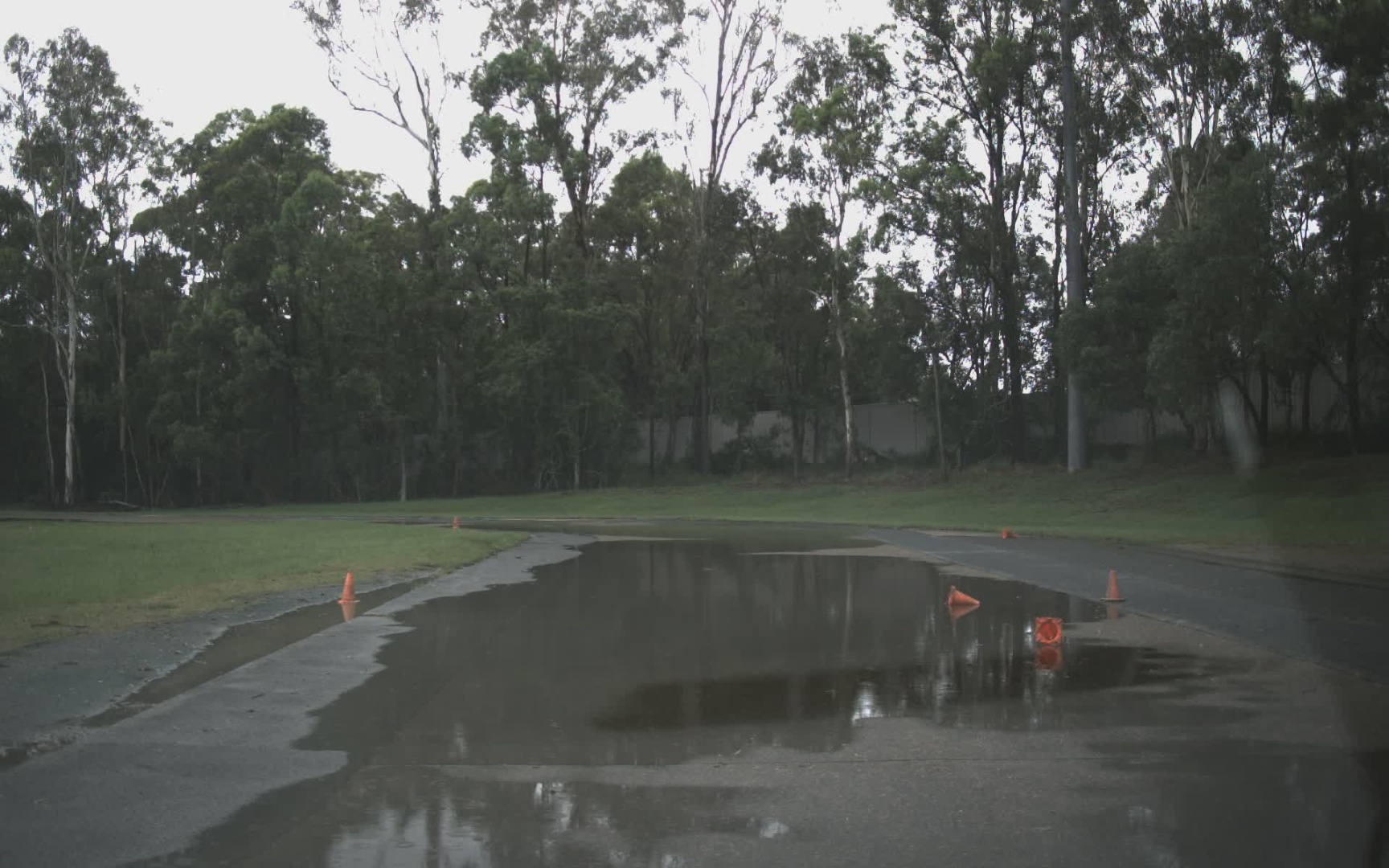} &
        \includegraphics[width=0.24\linewidth]{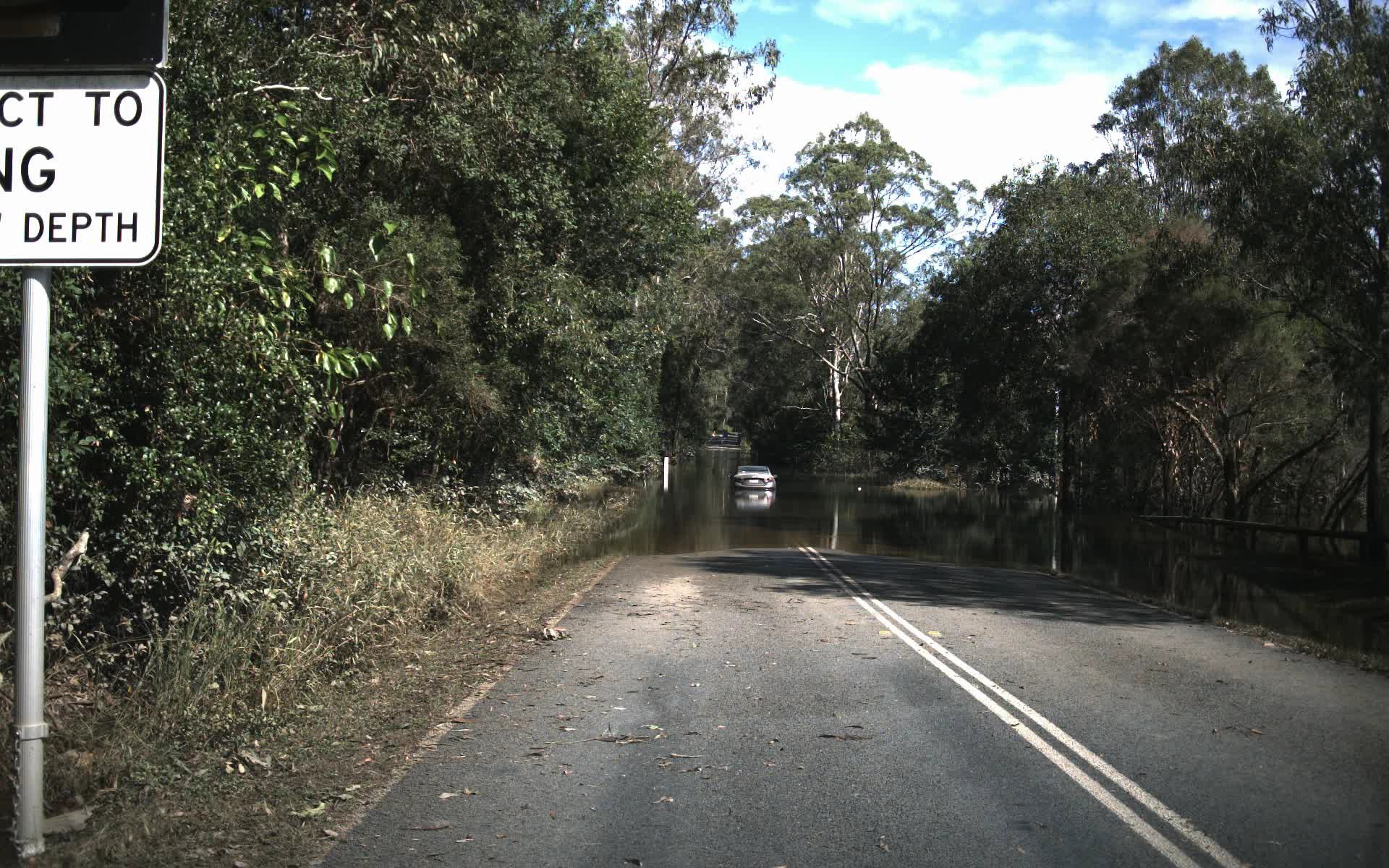} &  
        \includegraphics[width=0.24\linewidth]{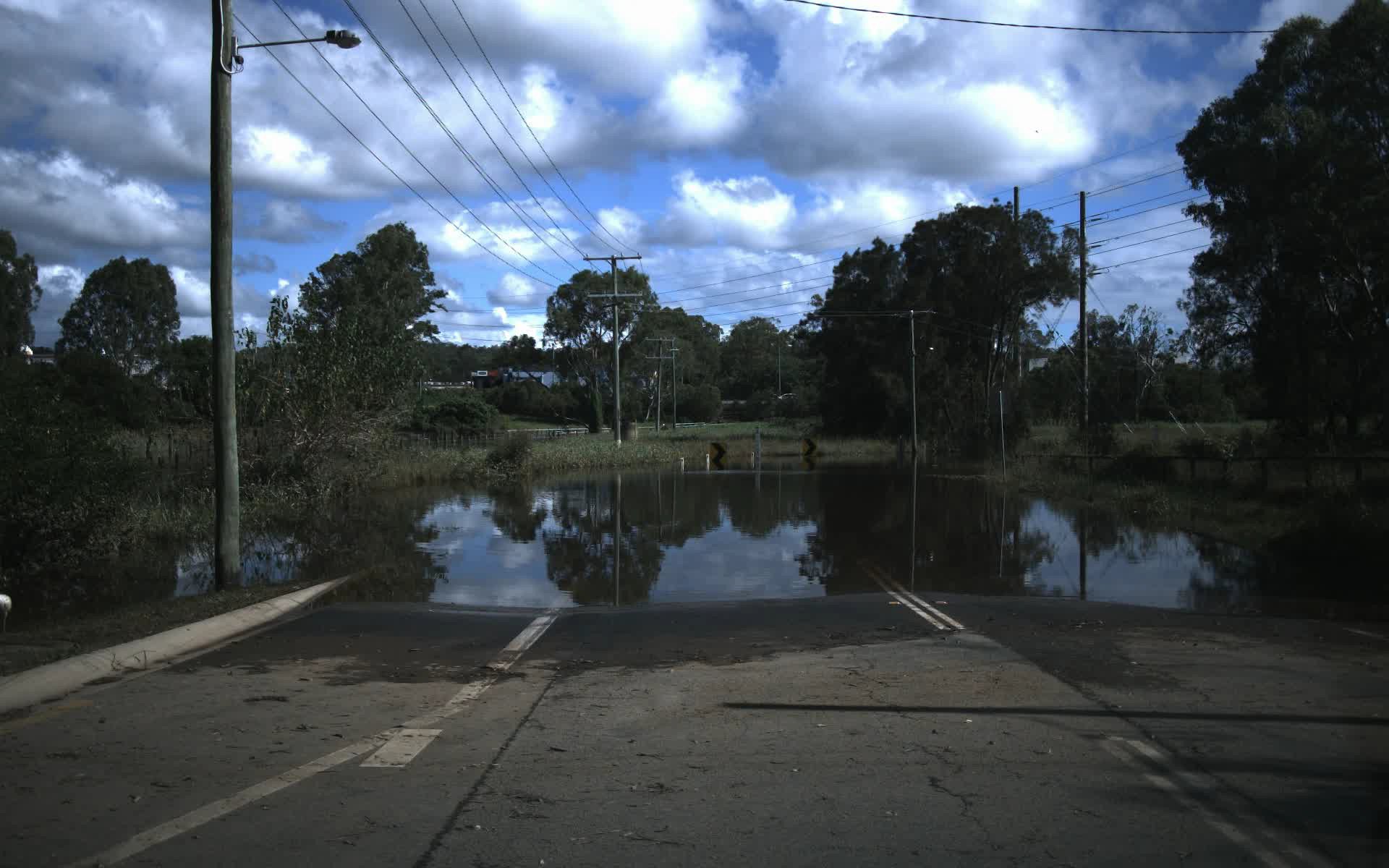} & 
        \includegraphics[width=0.24\linewidth]{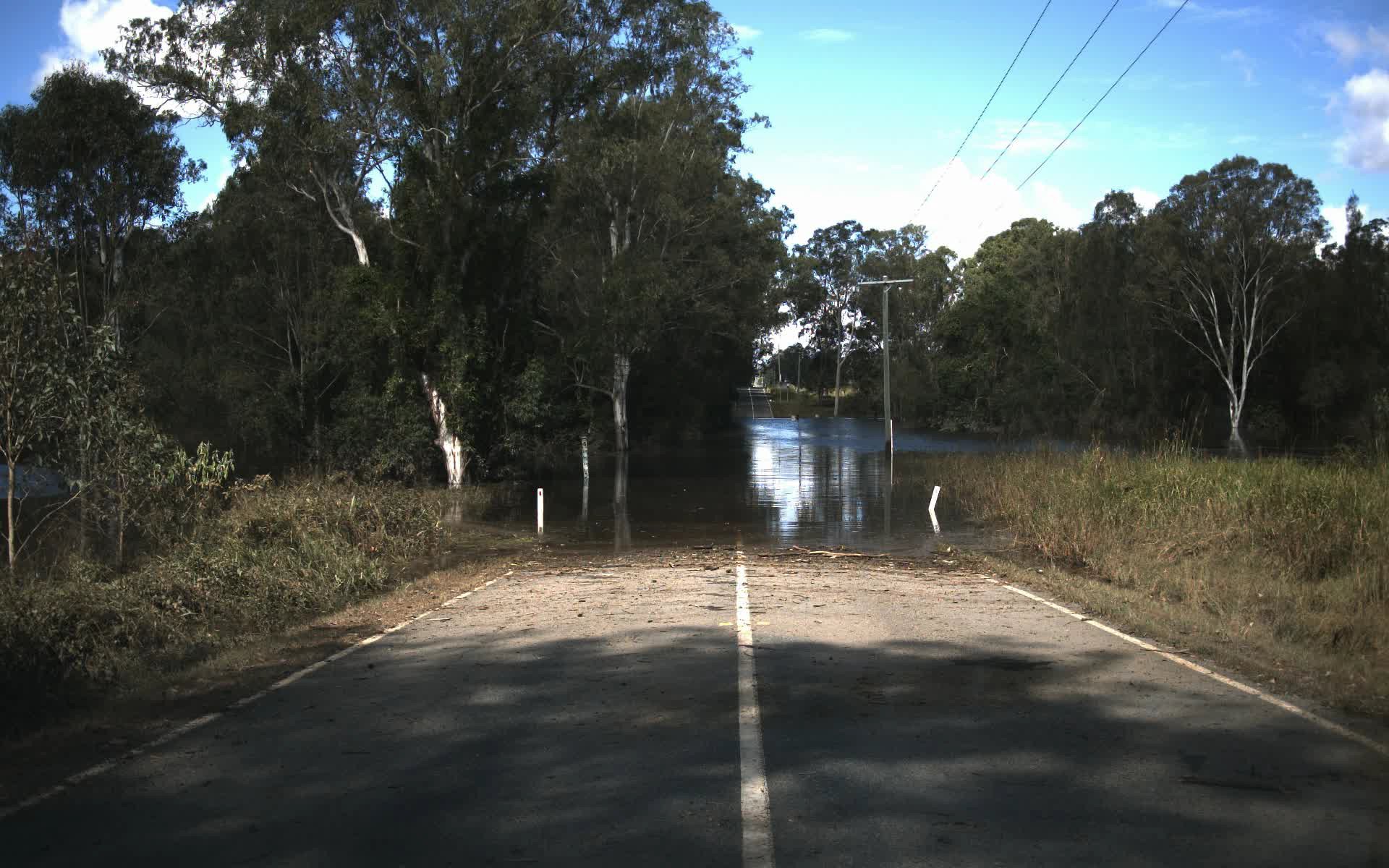}
    \end{tabular}
    \caption{Data across the five separate locations are unique and varied to encourage the development of more robust perception and localization methods. Mount Cotton includes puddle-like water hazards; Cambogan, Dairy Creek, and Holmview capture significant flooding events; and Pullenvale captures a running stream of water. The images above, from left to right, are from: Mount Cotton, Cambogan, Holmview, and Dairy Creek.}
    \label{fig:data-samples}
\end{figure*}
\setlength\tabcolsep{6pt}

\section{Data Collection}
\label{sec:data-collection}
The FRED dataset largely consists of data captured after a major flooding event in early 2025 around Brisbane, Australia. It was collected using a Renault Zoe equipped with a custom sensor payload to enable autonomous operation, referred to as the Zoe 2 (Figure~\ref{fig:zoe2}). The five locations visited for data collection will be referred to as Mount Cotton, Cambogan, Holmview, Pullenvale, and Dairy Creek. Sensor outputs from the autonomous vehicle were captured by manually driving the vehicle as close as possible to the respective water hazards while remaining safe. Figure~\ref{fig:data-samples} illustrates the unique water hazards and environment that each location captures with sample images from four of the locations.

\begin{table}
	\small\sf\centering
	\caption{Zoe 2 Sensor specifications.}
	
	\begin{tabular}{ll}
		\toprule
		Sensor & Specifications\\
		\midrule
		
		Geoflex RTK GNSS & $\bullet$ 1-3cm accuracy $^\dagger$ \\[1mm]
        
        iXblue ATLANS-C IMU & $\bullet$ 3.5-5cm positional RTK accuracy $^\dagger$\\
        & $\bullet$ 0.02$^\circ$ heading RTK accuracy \\
        & $\bullet$ 0.008$^\circ$ roll/pitch RTK accuracy \\
		& $\bullet$ 200 Hz \\[1mm]
        
		FLIR Blackfly Camera & $\bullet$ 1920x1200 (2.3 MP) \\
		(BFLY-U3-23S6C-C) & $\bullet$ 49.1$^\circ$ HFOV x 31.9$^\circ$ VFOV \\
		  & $\bullet$ 30 Hz \\ [1mm]
		
		Ouster OS1-64 & $\bullet$ 64 beams \\
		  & $\bullet$ 0.7$^\circ$ vertical resolution \\
		  & $\bullet$ 0.35$^\circ$ horizontal resolution \\
		& $\bullet$ 360$^\circ$ HFOV x $\pm$ 22.5$^\circ$ VFOV  \\
		& $\bullet$ 45m @ >90\% detection \\ & \ \ 
        \ (10\% reflectivity) \\
		& $\bullet$ $\sim$ 0.65M points/s \\
		& $\bullet$ 10 Hz \\ [1mm]
		\bottomrule
	\end{tabular}
	\label{tab:sensor-specs}
\end{table}

After sufficient time (months) for the flooding to subside and possible damage to be removed/repaired, each location was revisited and the sensor stack was captured on a pass through of the same locations without any water hazards. These `dry' runs of each location also recorded data through and passed where the water hazards were previously found to allow maps of the location to be built.

\section{Sensors}
\label{sec:sensors}
The Zoe 2 is equipped with a suite of sensors that allows fully autonomous operation. For the purpose of the FRED dataset, the sensors captured include front and rear FLIR Blackfly cameras, an Ouster OS1-64 LiDAR, and an iXblue ATLANS-C IMU corrected by a Geoflex RTK GNSS.
A summary of the sensor specifications can be found in Table~\ref{tab:sensor-specs}.

\section{Dataset Format}
\label{sec:dataset-format}
\subsection{Data Organization}
\begin{figure}[t]
    \centering
    \includegraphics[width=\linewidth]{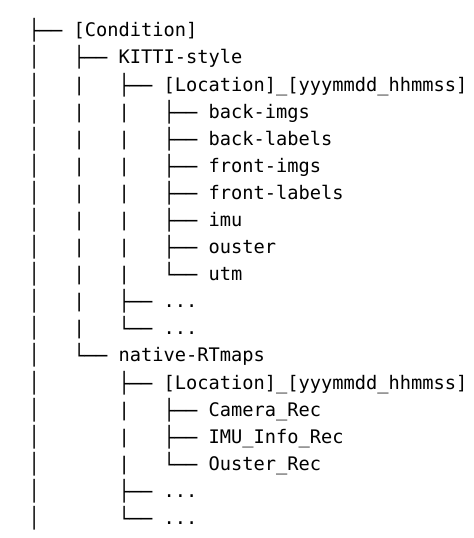}
    \caption{The FRED dataset separates sequences based on their condition (i.e. flooded or dry), location, and time of recording. In addition, each sequence is provided in both the native RT-Maps recording format and a KITTI-style format.}
    \label{fig:structure_tree}
\end{figure}

The FRED dataset is organized into individual sequences of data, each distinguished by four factors: the condition of the road, the location, the date and time of recording, and the format of the data. For this work, the condition of the road indicates whether it is `dry' or `flooded', the location refers to one of the five locations mentioned previously, the date and time indicate when the recording was started, and the data format conveys whether it is stored in the `native' format or the `KITTI-style' format. The date and time for each sequence are recorded using the year, month, day (yyyymmdd), and hour, minutes, seconds (hhmmss) formats, respectively. Figure~\ref{fig:structure_tree} shows the structure used to organize sequences in the dataset.

The Zoe 2 vehicle collects and records data using the RTMaps software; however, this software requires a licence and therefore is not freely accessible to researchers. Accordingly, sequences in FRED are provided in both the native RTMaps format for direct playback \textit{and} a KITTI-style format to align with conventions typically used in autonomous vehicle datasets for research.

\subsection{Native RTMaps Format}
RTMaps records data streams using four key files: a recording/metadata file (.rec), an index file (.idx), an identifier file (.idy), and the actual data stream. The recording file contains a human-readable and editable log of information, such as headers, session information, and references to data streams, which enables time-synchronised replays of the sequence. The index file helps provide `random' access to different parts of the recording during replay, which enables jumping to the desired timestamp/s in a recording. The identifier file contains information on how each data stream was recorded and how it should be decoded. Finally, each data stream is contained within a separate file using an appropriate file type. For example, camera data is stored in video format, LiDAR data is stored within a `stream 8' binary data stream file (.s8), and IMU data is stored in a text file.

In the FRED dataset, sequences in the native RTMaps format contain data streams from the front and rear cameras, the $360^{\circ}$ roof-mounted LiDAR, and the GNSS-corrected IMU. These data streams are stored within separate sub-directories within each sequence's parent directory and include their own recording, index, and identifier files. This enables playback of individual or selected data streams within the RTMaps software.

\subsection{KITTI-Style Format}
The KITTI-style format used within the FRED dataset aligns with the conventions commonly used in other autonomous vehicle datasets. In this format, data is sampled from the original recording at $\approx$10Hz and stored using the corresponding timestamps. The data from each sensor is stored within a separate sub-directory. Image data from cameras is stored in PNG format, LiDAR point clouds from each timestamp are stored in binary (.bin) files, and information from the GNSS and IMU is stored in text files. The data was not time synchronised during sampling; however, timestamps were created from a centralised computer and therefore can be used for synchronisation and/or alignment of the data for projection/fusion. Images have been anonymized using the updated implementation of Understand-AI's anonymizer\footnote{\url{https://github.com/fusionportable/Anonymizer}}.

LiDAR point clouds contain four data fields, $[x, y, z, i]$, which encode the position of a returned point with respect to the LiDAR's position on the vehicle, and the intensity of the returned signal. The intensity of each beam is described by a measurement of the returned photons normalised between $0$ and $255$. Points/measurements without a valid return signal are recorded in the point cloud with $x, y, z$ and intensity values of 0. This results in all point clouds containing $65,536$ points in identical order with respect to beam/ring ID and azimuth angle. The raw points are not in exact order for directly decoding into a structured `range image' type format, but the development kit (Section~\ref{sec:development-kit}) provides the configuration parameters required to destagger point clouds into this type of structured format. The point clouds have not been motion-corrected, but this can be accomplished using the vehicle position, speed, and yaw rate data provided by the IMU.

IMU data is decoded into text files containing latitude, longitude, altitude, the vehicle's forward velocity, i.e. speed (m/s), and the vehicle's yaw rate (rad/s). The GNSS position is recorded in a separate text file containing UTM coordinates.
\section{Calibration}
\label{sec:calibration}

\begin{figure}
    \centering
    \includegraphics[width=\linewidth]{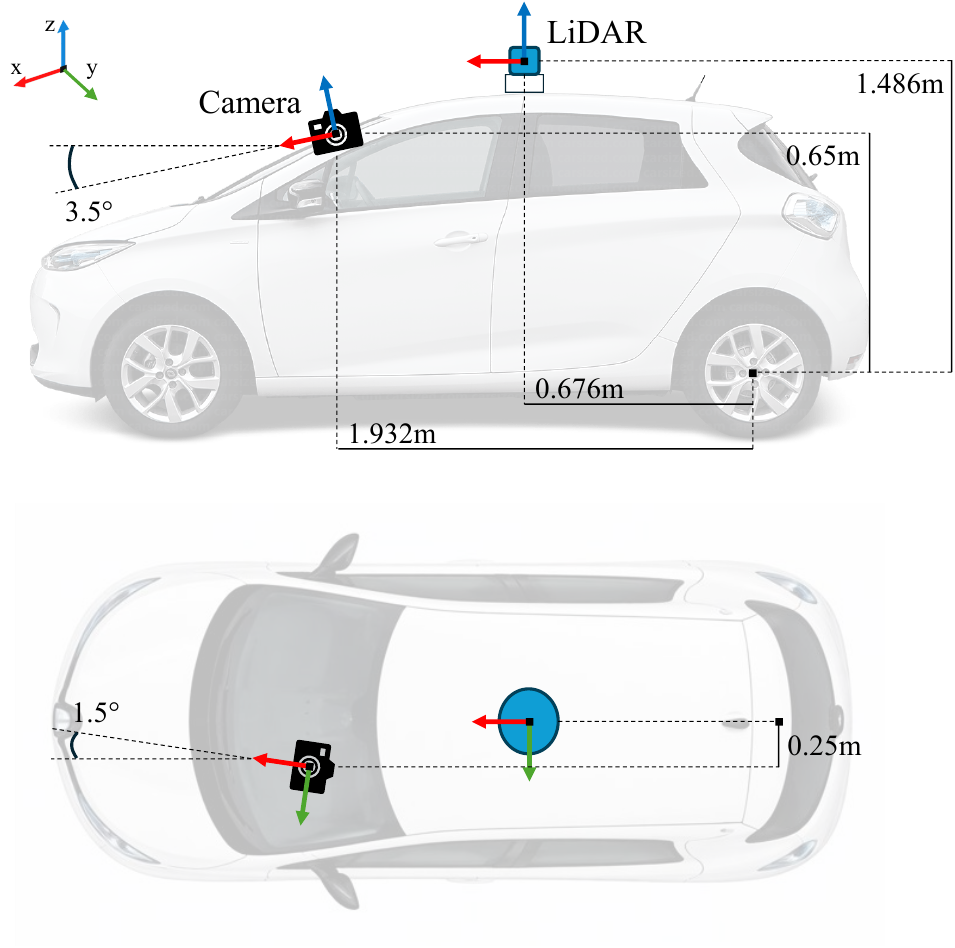}
    \caption{Sensors on the Zoe 2 that are relevant to the FRED dataset include a 360\textdegree Ouster LiDAR, and a front-facing RGB camera. The schematic above demonstrates how these sensors are positioned on the vehicle. The point of origin is centred over the rear axle where the IMU is positioned.}
    \label{fig:sensorExtrinsics}
\end{figure}

\subsection{Sensor Extrinsics}
The extrinsic measurements for each sensor are important for visualisation and data fusion. Figure~\ref{fig:sensorExtrinsics} provides the translation and rotation transformations between the LiDAR, front camera, and the common reference point. The centre of the car over the rear axle is used as a common reference point for sensor transformations to align with the IMU position. The development kit provides calibration files containing the necessary transformation matrices for the FRED dataset.

\subsection{Camera Intrinsics}
In addition to extrinsics used for sensor transformation matrices, camera \textit{intrinsics} are also required to accurately project LiDAR points onto images for sensor fusion. This includes focal lengths ($f_x$ and $f_y$), principal point coordinates ($c_x$ and $c_y$), and any distortion or skew coefficients. Using the manufacturers documentation for the FLIR Blackfly USB3 camera, images are captured with square pixels with focal length $f_x = f_y = 170.648$, and a principal point at $(c_x, c_y) = (960, 600)$. The FRED dataset provides rectified images, so the distortion and skew coefficients can be set to $0$.

\section{Annotations}
\label{sec:annotations}
\subsection{Image Annotations}
\begin{figure}
    \centering
    \includegraphics[width=\linewidth]{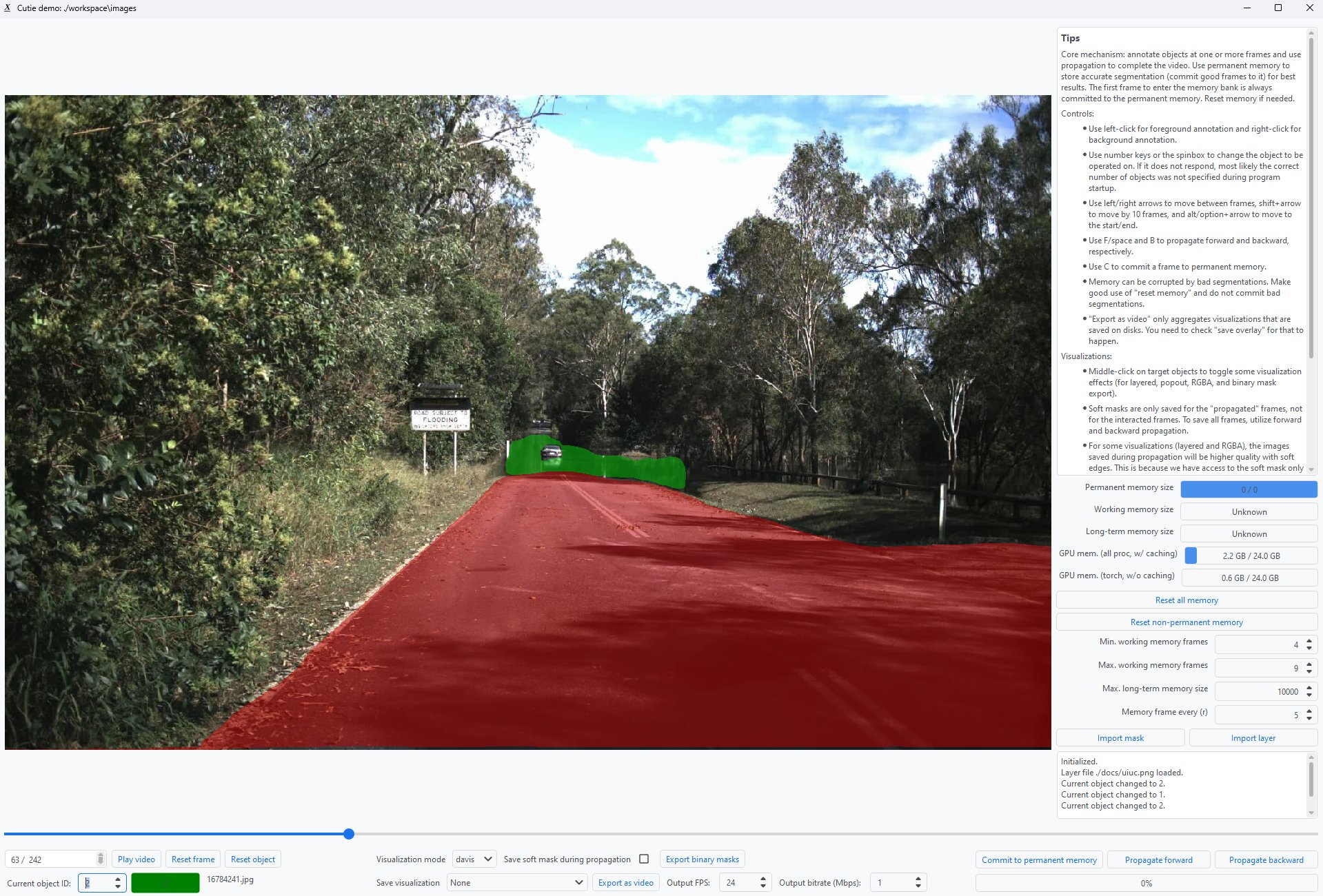}
    \caption{The Cutie image labelling tool provides an intuitive graphical user interface for annotating images. Sequences in the FRED dataset were annotated with red polygons for the road class and green polygons for water hazards.}
    \label{fig:imgAnnotation}
\end{figure}

To encourage research into the detection of water hazards in autonomous vehicles, the FRED dataset provides semantic annotations for images taken from the front camera across all `flooded' sequences. Annotations were created using three semantic classes: `water hazard', 'road', and 'other'. To improve the efficiency of manually creating annotations, the `Cutie' video object segmentation network\footnote{\url{https://github.com/hkchengrex/Cutie}}~\cite{cheng2024putting} was utilised to perform label propagation between frames (Figure~\ref{fig:imgAnnotation}). Any annotations created through this propagation tool were then manually checked and adjusted to ensure accurate labelling. Annotations/labels are provided as separate PNG image files named using the corresponding image timestamp.

\subsection{Point Cloud Annotations}
\begin{figure}
    \centering
    \includegraphics[width=\linewidth]{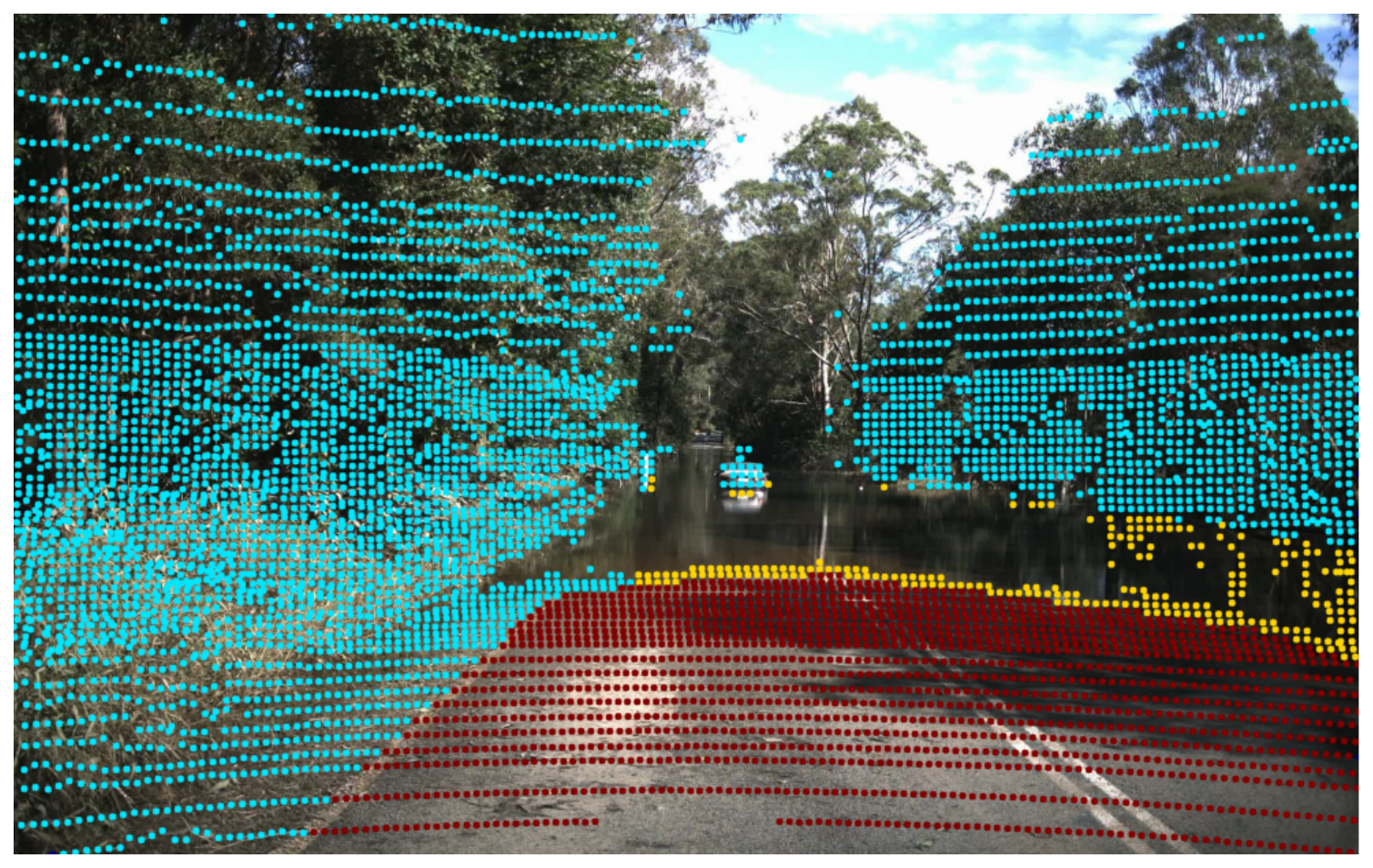}
    \caption{Point clouds can be annotated by projecting points and adopting labels from images at the corresponding time step. }
    \label{fig:pointcloudAnnotation}
\end{figure}

One of that factors that makes detecting water hazards challenging is that LiDARs often do not reliably return points across the majority of the water. For this reason, the FRED dataset does not explicitly provide semantic labels for the recorded point clouds, however, the development kit provides tools which can be used to create annotations. Points can be labelled by first projecting them onto the image plane and then adopting the image label at the corresponding location (Figure~\ref{fig:pointcloudAnnotation}). This method cannot be used to label all points within a point cloud, but it can be used to provide labels to points that are critical for detecting water hazards on the road.
\section{Development Kit}
\label{sec:development-kit}
\begin{figure*}
    \centering
    \begin{tabular}{ccc}
        \includegraphics[width=0.32\linewidth]{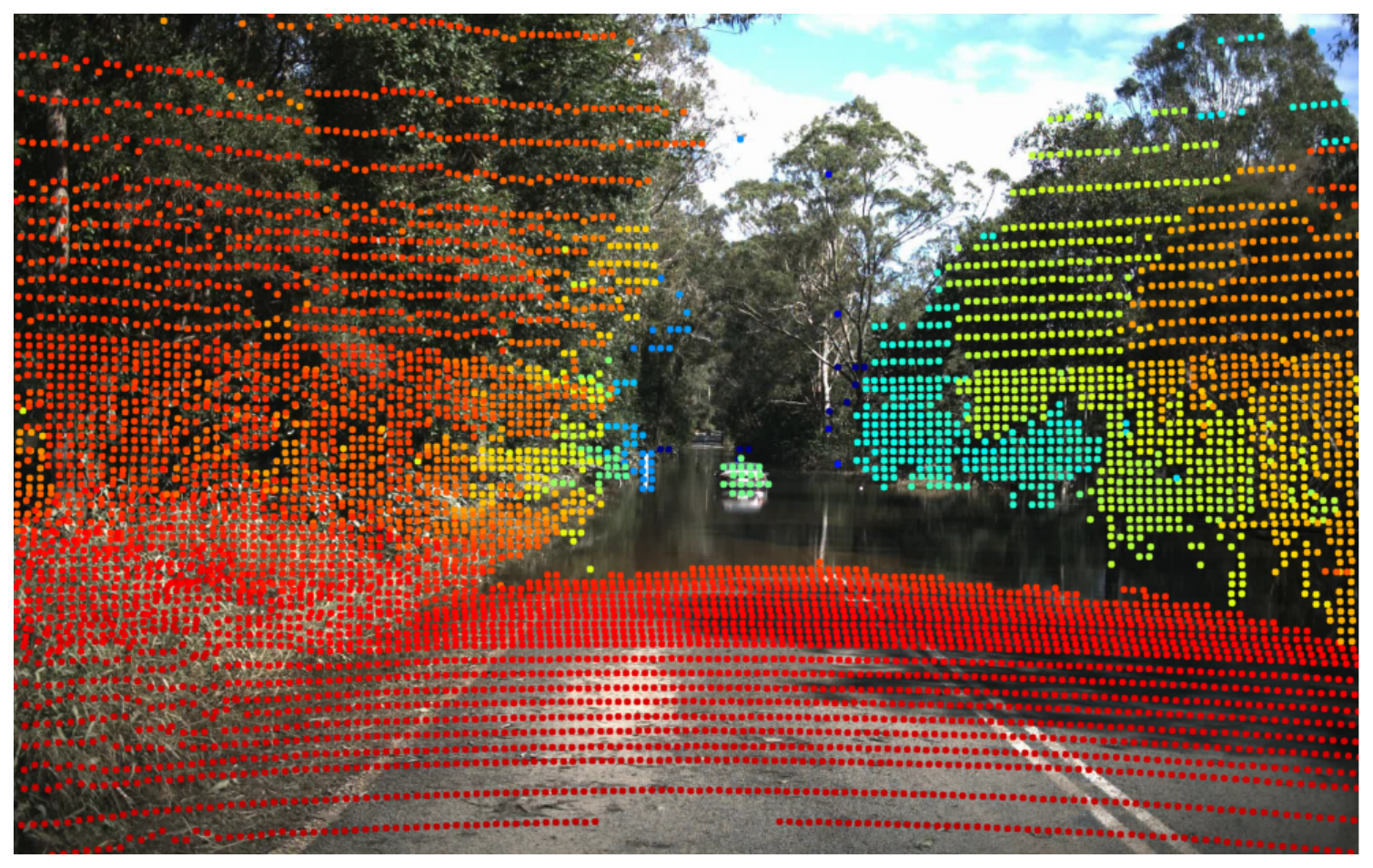} &
        \includegraphics[width=0.32\linewidth]{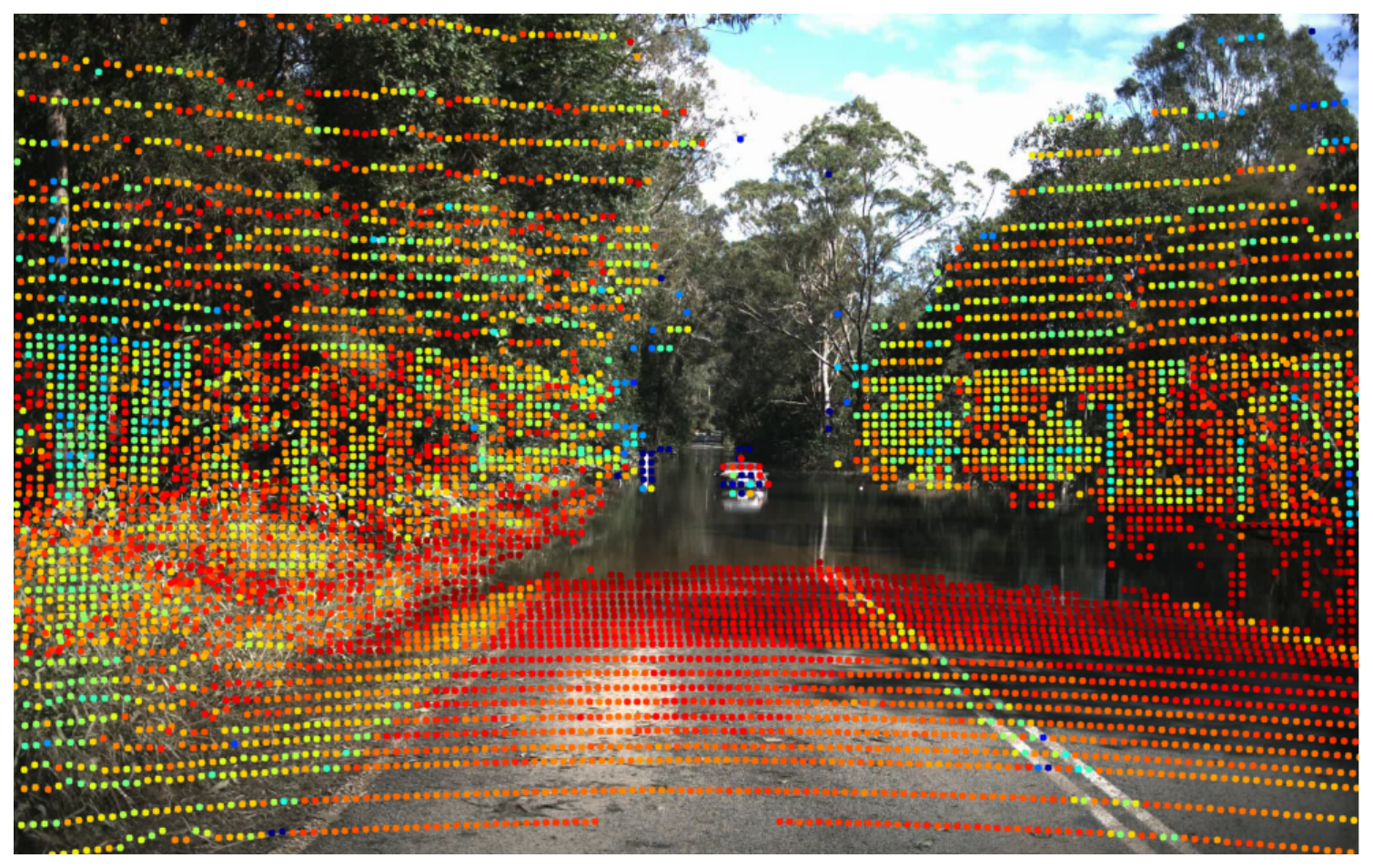} &
        \includegraphics[width=0.32\linewidth]{figures/labelled_pointcloud.pdf}
    \end{tabular}
    \caption{The FRED software development kit provides three ways to colour LiDAR points that are projected onto an image. \\ \textbf{Left:} Using distance/range calculations. \textbf{Middle:} Using intensity measurements. \textbf{Right:} Using semantic labels.}
    \label{fig:projectpoints}
\end{figure*}

\begin{figure*}
    \centering
    \includegraphics[width=\linewidth]{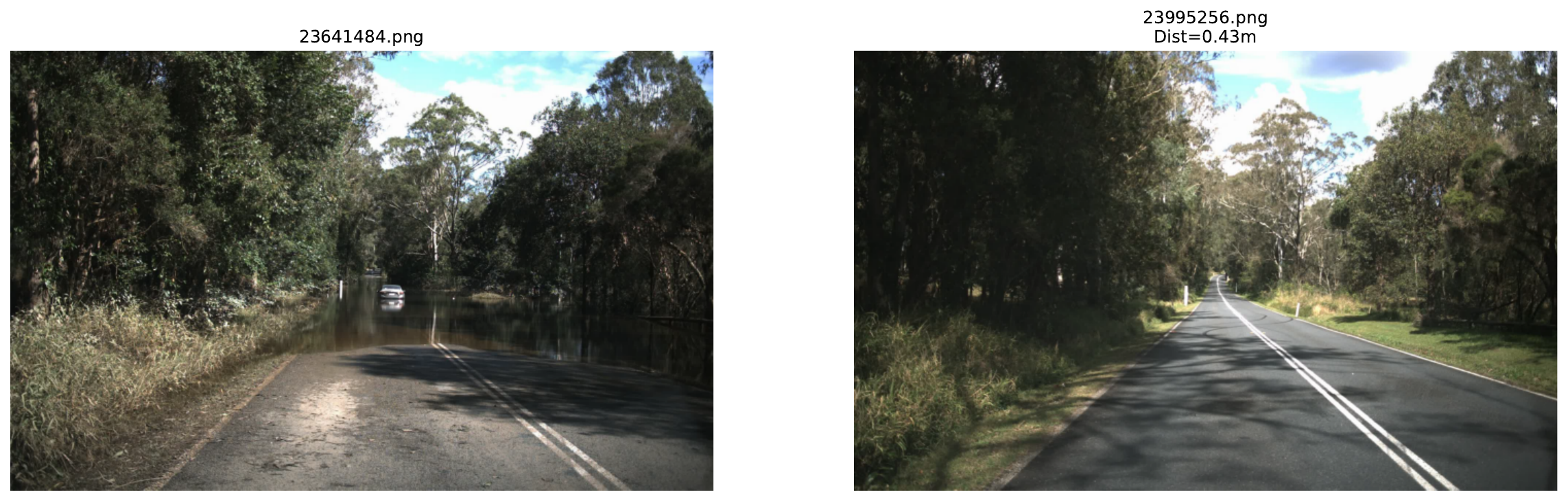}
    \caption{The FRED dataset includes sequences from the same locations captured under both flooded and dry conditions. The software development kit includes tools for searching across sequences for images captured from the same location.}
    \label{fig:localization}
\end{figure*}

\begin{figure}
    \centering
    \includegraphics[width=\linewidth]{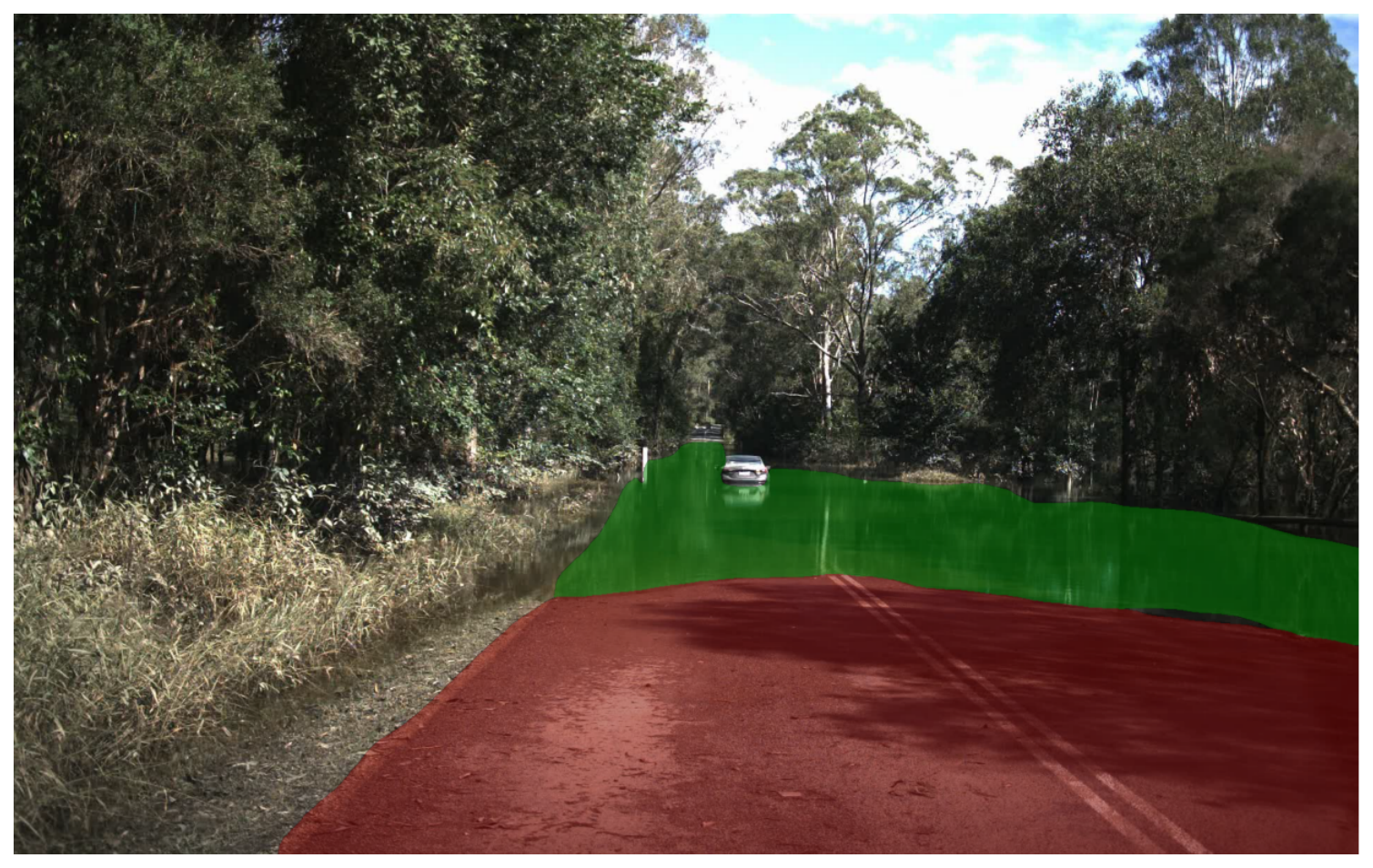}
    \caption{Images in the FRED dataset are provided with semantic labels. The annotations include a road class (red) and a water hazard class (green).}
    \label{fig:semanticLabels}
\end{figure}

\begin{figure}
    \centering
    \includegraphics[width=\linewidth]{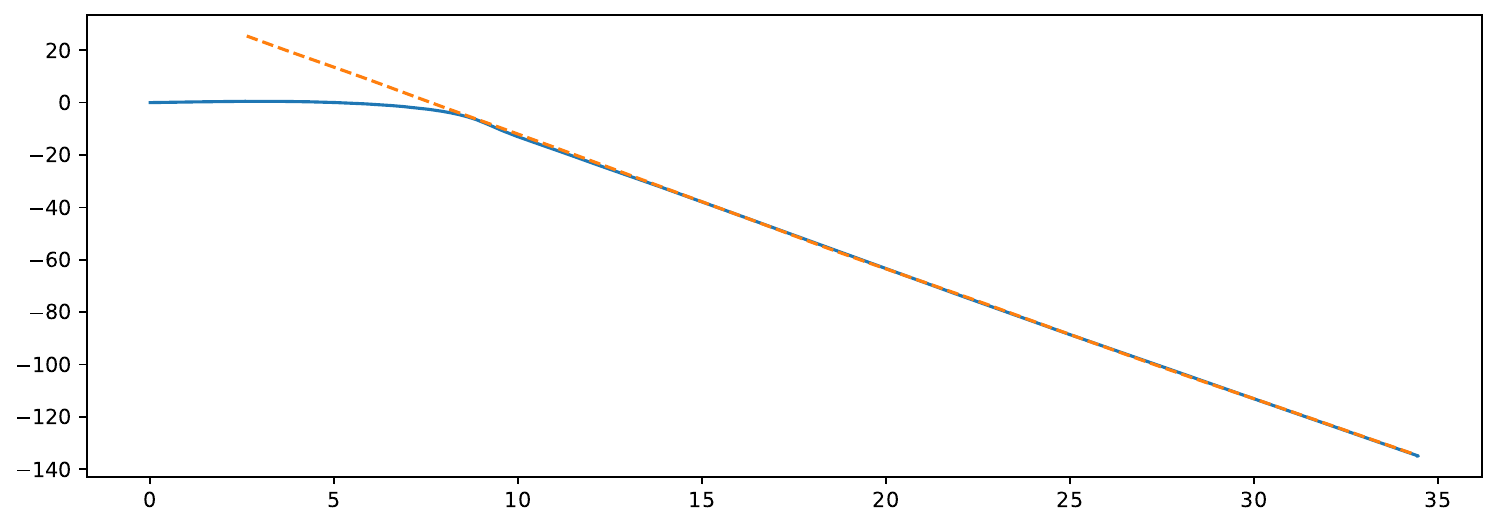}
    \caption{The FRED software development kit includes a tool for plotting the UTM trajectory of two sequences to validate their alignment. Here, the flooded Cambogan sequence is plotted in blue and the dry Cambogan `20250812 122339' sequence is plotted in orange.}
    \label{fig:trajPlot}
\end{figure}

\begin{figure}
    \centering
    \includegraphics[width=\linewidth]{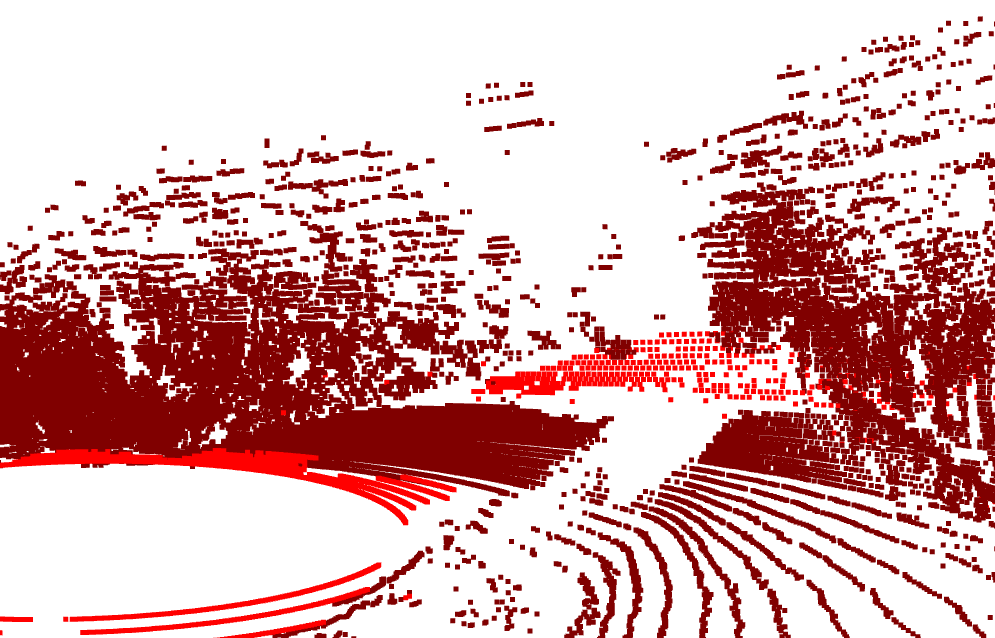}
    \caption{Sequences in the FRED dataset that are captured in flooded conditions typically contain `missing' points where water hazards prevent a returned signal. The development kit includes an experimental tool for projecting points onto the ground plane in these empty regions. The new projected points are plotted in bright red, and the original points are plotted in darker red.}
    \label{fig:pointcloudCompletion}
\end{figure}

To encourage and foster research into detecting water hazards, we provide a Python-based development kit\footnote{\scriptsize{\url{https://github.com/AVR3-Training-Centre/python-FRED}}} for the FRED dataset. The development kit includes tools commonly used in segmentation and localisation tasks for loading, manipulating, visualising, and evaluating data. In addition, it provides calibration and configuration files to enable the use of sensor fusion approaches. The tools provided include Python scripts for projecting point clouds onto corresponding images (Figure~\ref{fig:projectpoints}); visualising semantic annotations (Figure~\ref{fig:semanticLabels}); displaying corresponding images from different sequences for visual localisation (Figure~\ref{fig:localization}); plotting trajectories from corresponding sequences (Figure~\ref{fig:trajPlot}); and point cloud completion by projecting missing points onto the ground plane (Figure~\ref{fig:pointcloudCompletion}). The development kit will continue to be updated to increase its function and utility.

\setlength\tabcolsep{1.5pt}
\setlength\tabcolsep{6pt}

\section{Benchmark Metrics and Evaluation}
\label{sec:benchmark-metrics}
To complement the proposed FRED dataset, we provide evaluations of some common tasks for autonomous platforms to benchmark current methods in these flooded environments. Specifically, we evaluate a range of methods across image-based semantic segmentation and Visual Place Recognition (VPR). 

\subsection{Image-Based Semantic Segmentation}
\subsubsection{Overview}\ \\
Semantic segmentation is the task of classifying parts of an image into different object categories. This is critical for real-time scene understanding and perception in autonomous vehicles and robot platforms. Typically, modern approaches for semantic segmentation of images employ either convolutional neural networks (CNNs) or Vision Transformers (ViTs) to produce a set of pixel-wise semantic labels that can be used for subsequent tasks, such as object avoidance (\cite{lateef2019survey, thisanke2023semantic}).

Segmentation networks often perform most accurately and robustly for objects that have consistent visual features (i.e. colour, shape, texture, etc.). Accordingly, water hazards have often proved challenging to accurately detect due to the large variation in appearances (\cite{rankin2006daytimeugv}). Some recent works have explored fine-tuning networks or the use of attention modules for detecting reflections on the surface of water hazards (\cite{han2018single, wang2019water}). However, research is limited and implementations of these methods are often not publicly released. Additionally, most existing works benchmark performance against non-public datasets or the Puddle 1000 dataset, which is no longer accessible\footnote{Due to the decommissioning of CloudStor.} and does not cover a wide range of scenarios. Given the danger that water hazards, such as flooded roads, pose to autonomous platforms, it is important to increase the number of supported datasets for improving detection performance and accurately evaluating the robustness of existing approaches.

\subsubsection{Metrics: Intersection Over Union}\ \\
The metric generally used in image-based semantic segmentation for evaluating the accuracy of models is intersection over union (IoU),  also known as the Jaccard Index. IoU measures the overlap between predicted and ground-truth regions of an image for a given class. It is calculated by dividing the overlapping area of these two regions by the total area covered by both:
\begin{equation}
\text{IoU}_c = \frac{|P_c \cap G_c|}{|P_c \cup G_c|}
\end{equation}

Where $P_c$ is the regions of an image that are predicted to belong to object $c$, and $G_c$ is the \textit{actual} regions of an image that belong to object $c$. An IoU of 1 (or 100\%) indicates a method that has correctly classified the object that every pixel in an image belongs to, and 0 indicates that no pixels were correctly classified.

When there is no ground truth and no predicted region in an image containing a particular class, the IoU metric becomes undefined due to division by zero. This can significantly affect the calculation of the \textit{Mean} IoU (mIoU) across an entire dataset. Generally, one of three strategies is adopted to account for this: (1) treat these images as having an IoU of 1, (2) treat these images as having an IoU of 0, or (3) remove these images from the calculation of mIoU. In the following evaluations, we treat instances where there is no water hazard/s on the road \textit{and} no predicted areas of water to have an IoU of 1. In the context of detecting water hazards during autonomous vehicle operation, it is important to also capture when a system has correctly identified that there is no water on the road.

\subsubsection{Experimental Setup}
\begin{figure}
    \centering
    \renewcommand\tabcolsep{3pt}
    \begin{tabular}{cc}
         \includegraphics[width=0.49\linewidth]{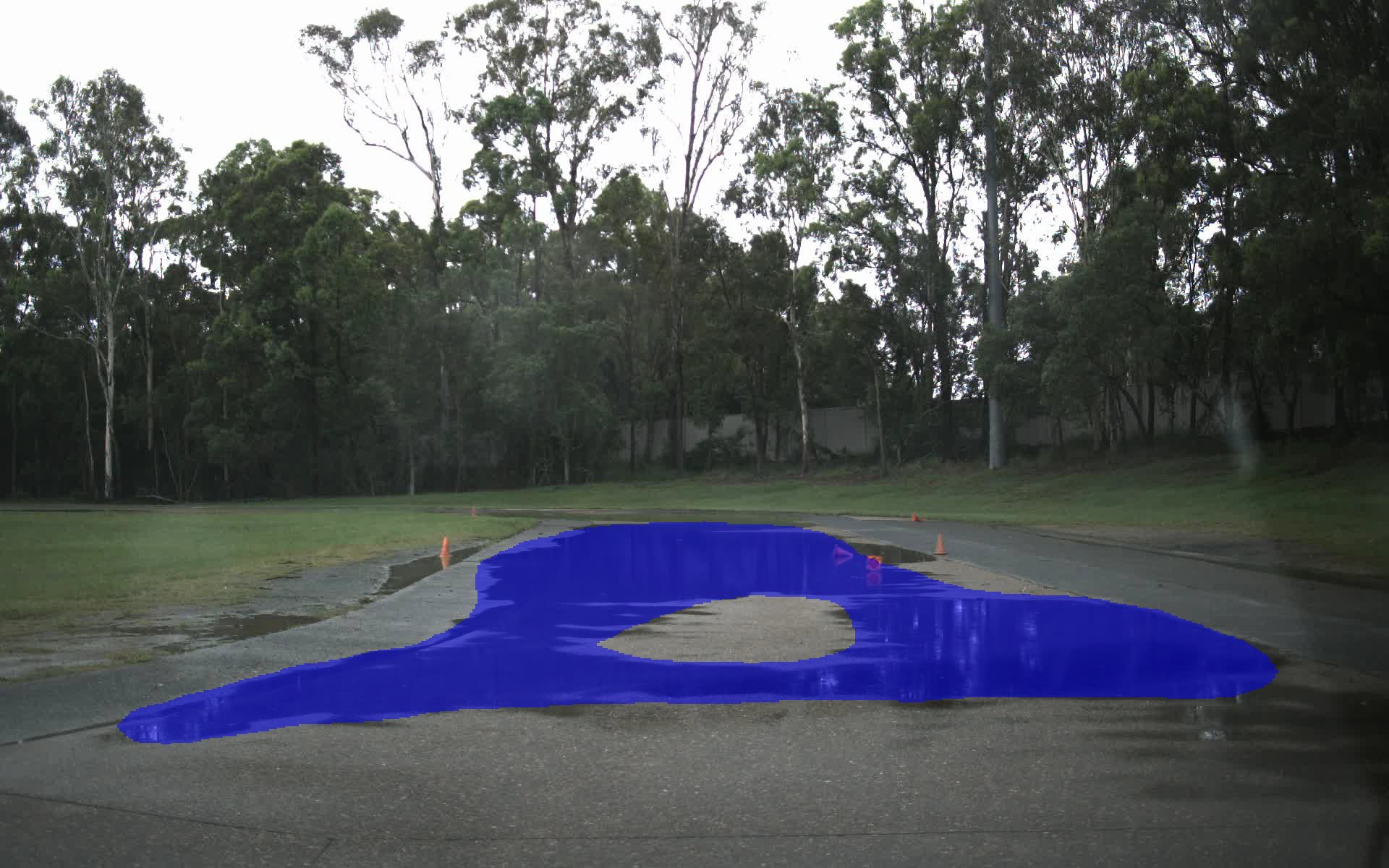}  &
         \includegraphics[width=0.49\linewidth]{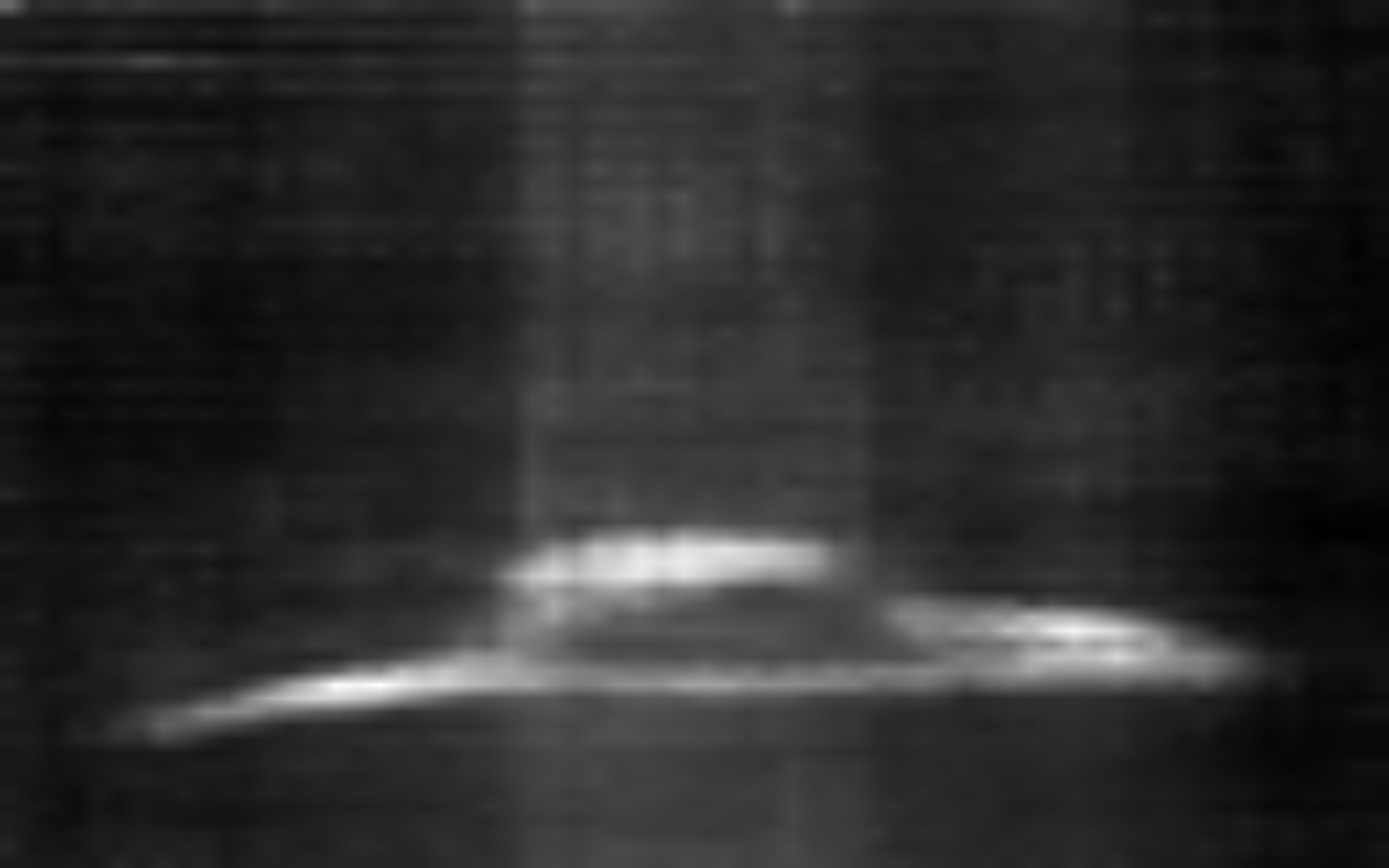} 
    \end{tabular}
    \caption{The reflection attention unit (RAU) enables neural networks to focus on learning features in regions of the image that contain reflections. \textbf{Left:} Segmentation results from training a Deeplab V3 segmentation network with an RAU layer on the Mount Cotton sequence. \textbf{Right:} The corresponding attention mask from the RAU layer.}
    \label{fig:rau}
\end{figure}

\ \\
To demonstrate the need for research on the detection of water hazards, we provide benchmark results using a selection of the most recent computer vision approaches for water detection. The Reflection Attention Unit (RAU) was a key development that enabled more targeted training of Convolutional Neural Networks (CNNs) for segmentation of water hazards (\cite{han2018single}). The original work implemented this attention unit inside an FCN-8 architecture to improve the detection of reflections on water surfaces. The original implementation was not provided with a trained model and the released code was found to contain errors. As a result, we provide an updated implementation that integrates the attention module into the more recent DeepLab-V3 architecture. The functionality of the attention module was validated by observing the attention layer outputs during training (Figure~\ref{fig:rau}). A model was trained with this architecture using the Mount Cotton `flooded' sequence from FRED and tested on the remaining flooded sequences.

There are also some computer vision tasks, such as terrain classification and flood monitoring, that are adjacent to the detection of flooded roads and could be useful for segmenting water hazards. We provide segmentation results using recent models from terrain classification and flood monitoring on the FRED dataset. GA-Nav is a semantic segmentation model trained for off-road terrain classification (\cite{ganav2022guan}). V-Flood is a model trained for urban flood detection and quantification using video (\cite{liang2023v}). Finally, we also include segmentation results a YOLOv8 model (\cite{duchieu2605032023flood}) trained on the water segmentation dataset from \cite{liang2020waternet}. The original model trained in \cite{liang2020waternet} is designed specifically for stationary cameras and therefore could not be directly used in this context.

\begin{table*}[t]
    \centering
    \scriptsize
    \caption{mIoU results for the segmentation of water hazards across datasets.}
    \label{tab:miou}
    \begin{tabular}{lccccc}
        \toprule
        \textbf{Method} & \textbf{Cambogan} & \textbf{Dairy Creek} & \textbf{Holmview} & \textbf{Mount Cotton} & \textbf{Pullenvale} \\
        \midrule
        Deeplab V3 RAU (\cite{han2018single}) & 31.06 & 6.93 & 18.82 & \textit{86.88} & 17.99 \\
        GA-Nav (RUGD) (\cite{ganav2022guan}) & 13.59 & 0.0 & 17.0 & 21.78 & 20.83 \\
        GA-Nav (RELLIS) (\cite{ganav2022guan}) & 0.56 & 5.75 & 2.75 & 23.0 & 7.62 \\
        V-Flood (\cite{liang2023v}) & 19.37 & 43.42 & 19.77 & 30.31 & 19.53 \\
        YOLOv8 (\cite{duchieu2605032023flood}) & 12.35 & 0.0 & 25.0 & 19.22 & 24.88 \\
        \bottomrule
    \end{tabular}
\end{table*}

\begin{figure*}
    \centering
    \renewcommand\tabcolsep{3pt}
    \begin{tabular}{cccc}
        \includegraphics[width=0.23\linewidth]{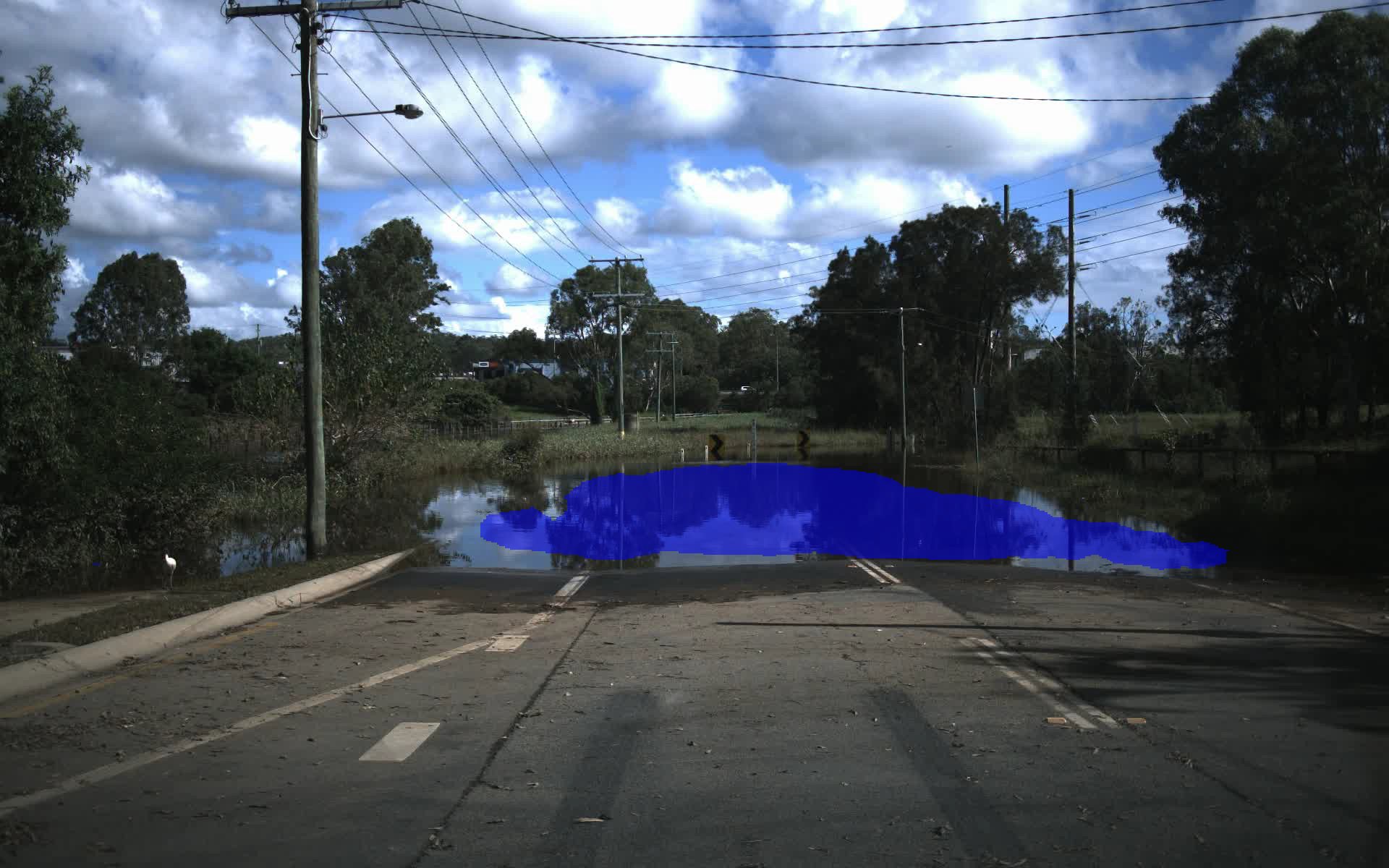} &  \includegraphics[width=0.23\linewidth]{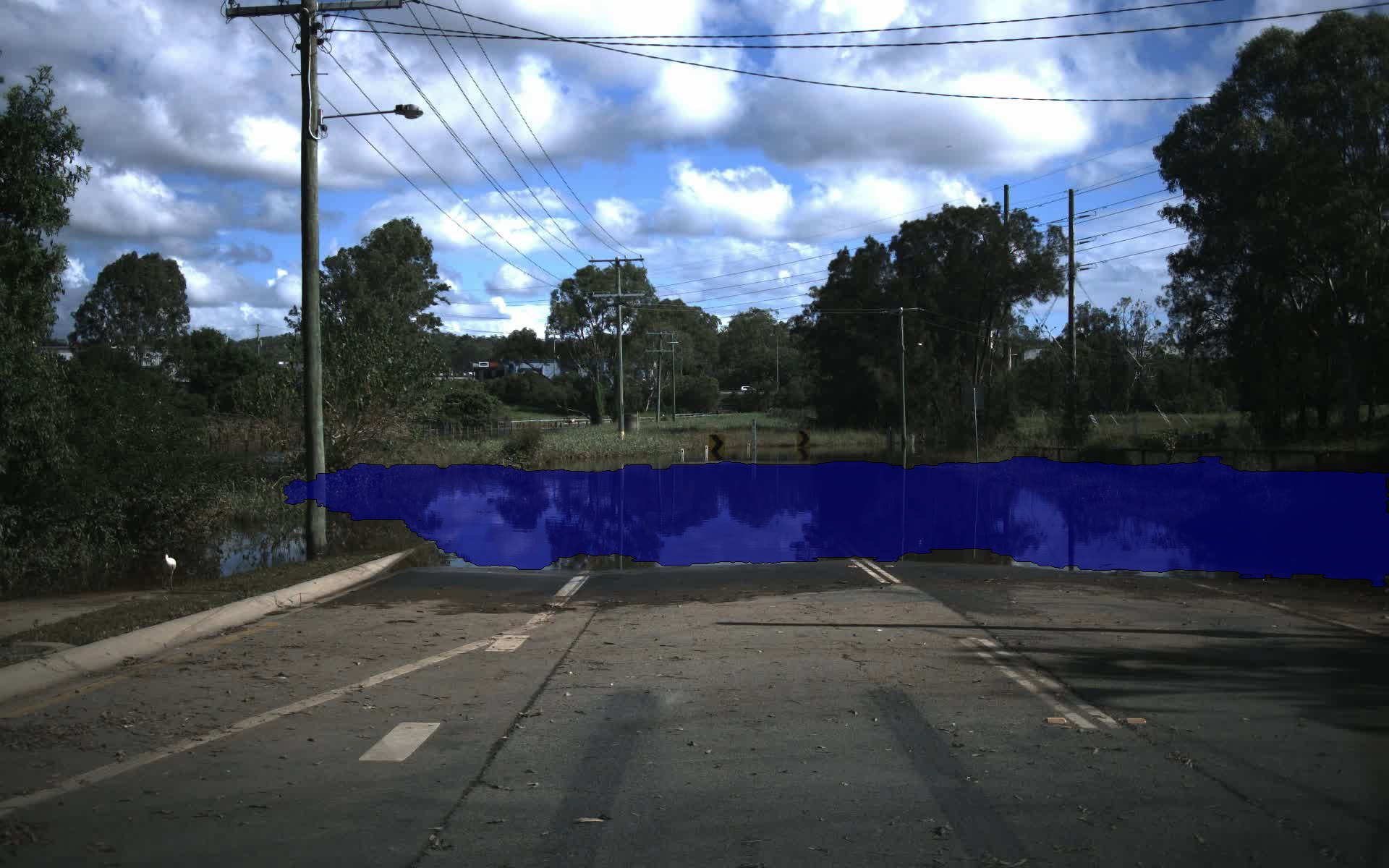} &
        \includegraphics[width=0.23\linewidth]{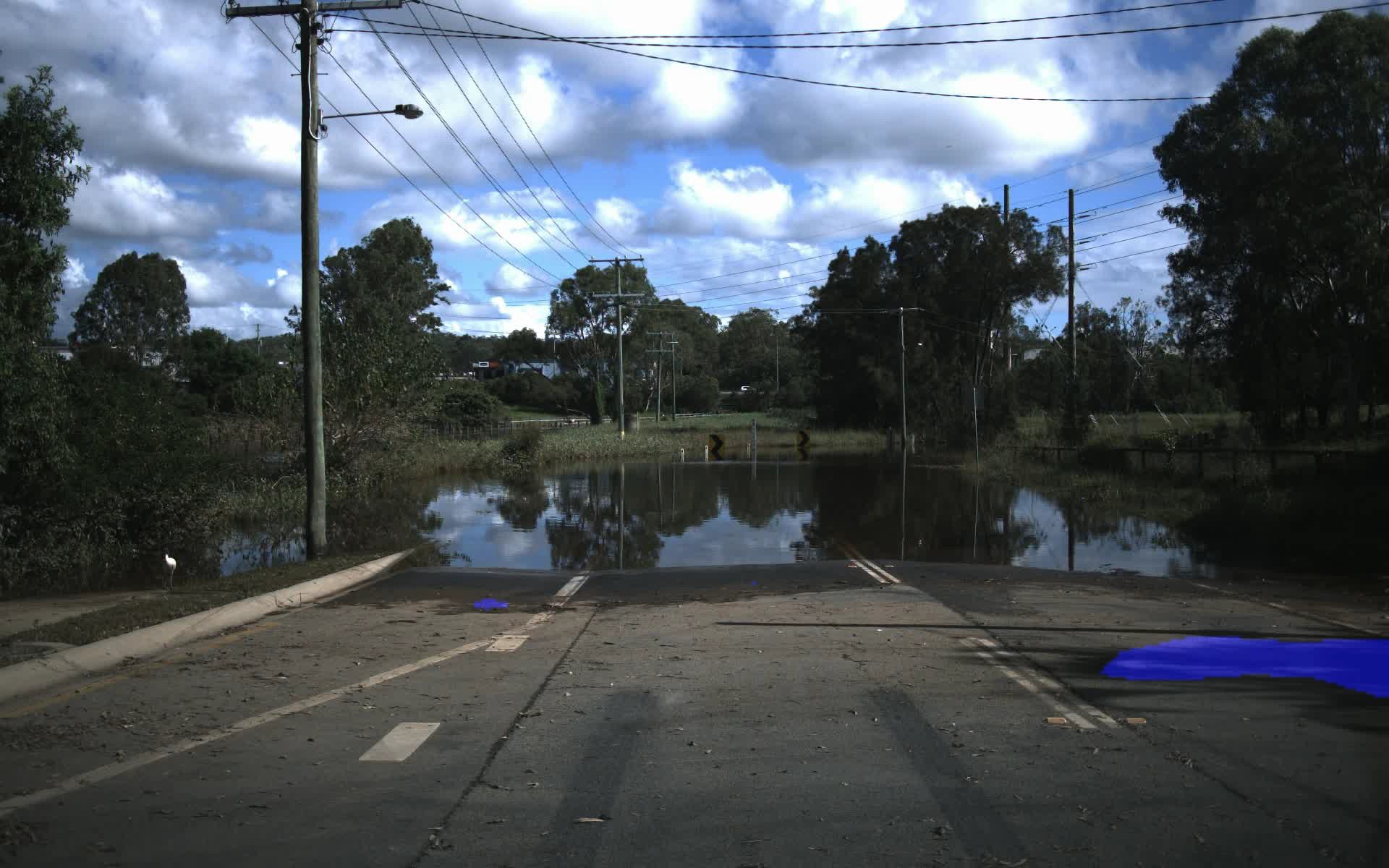} &  \includegraphics[width=0.23\linewidth]{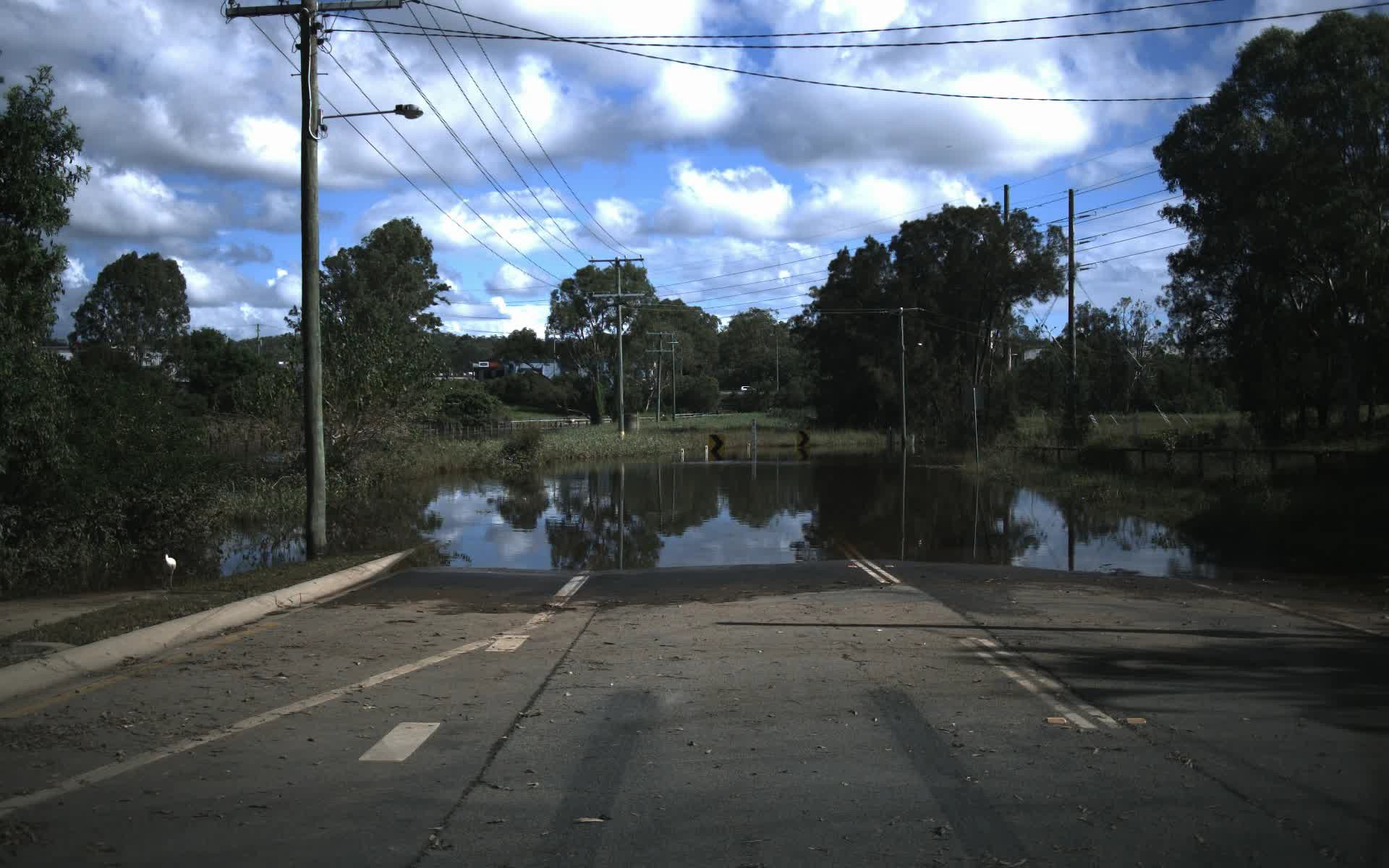} \\
        Deeplab V3 RAU & V-Flood & GA-Nav (RUGD) & YOLOv8
    \end{tabular}
    \caption{Results demonstrate that V-Flood and Deeplab V3 RAU show the most promise for water hazard segmentation. GA-Nav and YOLOv8 trained on the WaterNet dataset generalised poorly beyond their original applications.}
    \label{fig:segcomparison}
\end{figure*}

The implementation of GA-Nav is provided with models trained on both the RUGD (\cite{wigness2019rugd}) and RELLIS3D (\cite{jiang2020rellis3d}) off-road datasets. Both of these datasets include instances of water hazards, with dedicated semantic classes for segmentation. The GA-Nav model reduces the full set of semantic classes from these datasets to a subset of six higher level classes (smooth, rough, muddy/bumpy and forbidden terrains, as well as obstacles and background). The `forbidden terrain' class is used for the following evaluation because it encompasses the relevant water classes. For completeness, we provide results from models trained on both respective datasets.

\subsubsection{Results and Discussion}\ \\
Table~\ref{tab:miou} demonstrates that all of the existing methods for water hazard segmentation performed worse on the FRED dataset than datasets in their published results. This indicates that existing methods do not generalise well outside of the conditions/environments evaluated in their respective publications. V-Flood and Deeplab V3 RAU showed promising performance, while GA-Nav and YOLOv8 generally performed poorly on the FRED dataset (Figure~\ref{fig:segcomparison}). Analysis of individual results reveals the specific challenges faced in water hazard segmentation.

\begin{figure}
    \centering
    \renewcommand\tabcolsep{3pt}
    \begin{tabular}{cc}
        \includegraphics[width=0.49\linewidth]{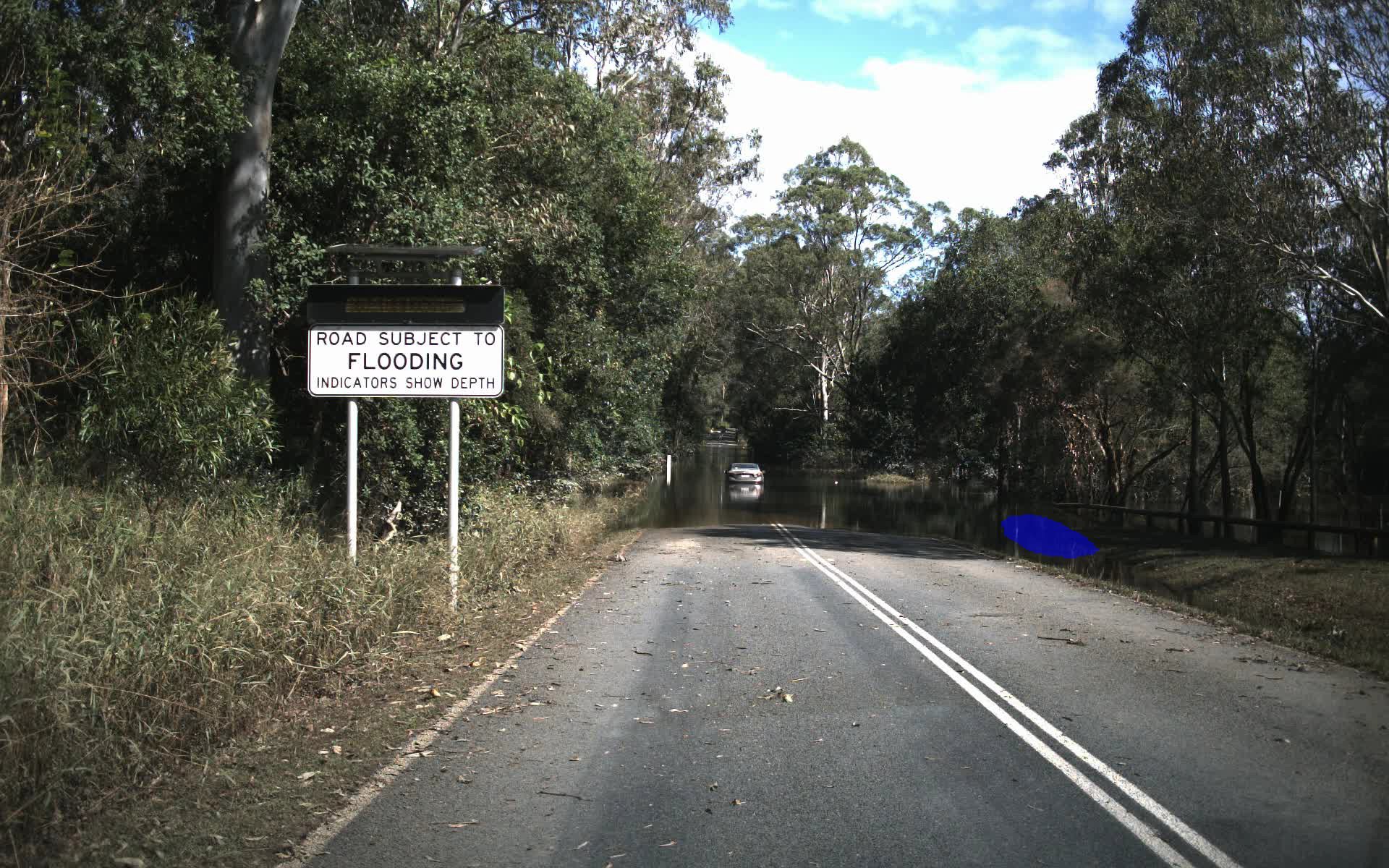} &
        \includegraphics[width=0.49\linewidth]{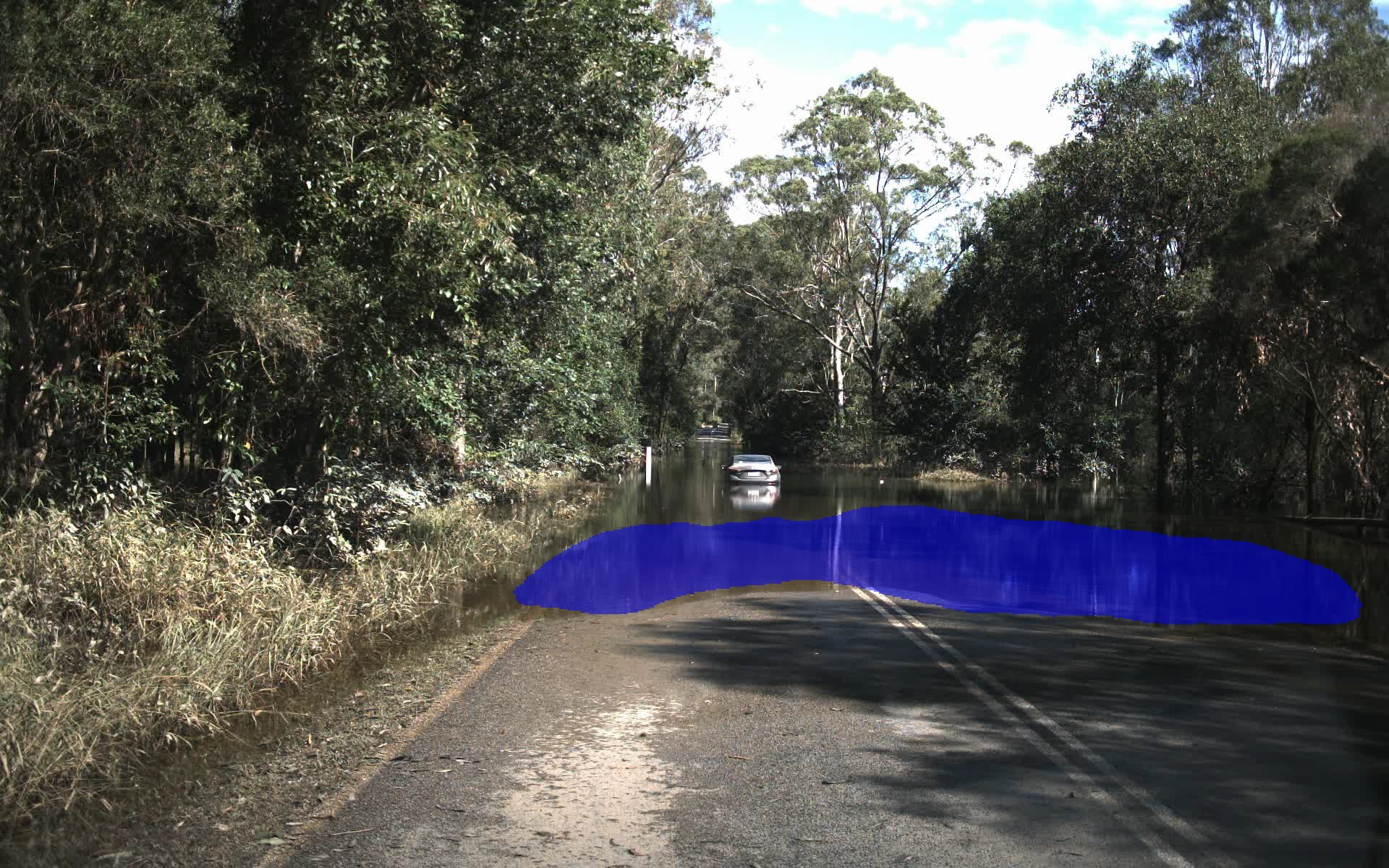}
    \end{tabular}
    \caption{One of the challenges for existing segmentation methods is detecting water hazards at a sufficient distance for autonomous vehicles to stop or avoid them. The above figures show poor segmentation by Deeplab V3 RAU at moderate distance (\textbf{Left}) but relatively good segmentation at close distance (\textbf{Right}) for the Cambogan sequence.}
    \label{fig:longDet}
\end{figure}

One particular challenge that was noticeable across all of these existing methods was detecting water hazards at longer distances. For autonomous vehicles, detecting flooded roads at longer ranges is important for ensuring the vehicle has sufficient distance to decelerate and avoid the hazard. Figure~\ref{fig:longDet} illustrates that Deeplab V3 RAU is unable to detect the majority of flooding on a road at a moderate distance, despite decent performance at closer distances.

\begin{figure}
    \centering
    \renewcommand\tabcolsep{3pt}
    \begin{tabular}{cc}
        \includegraphics[width=0.49\linewidth]{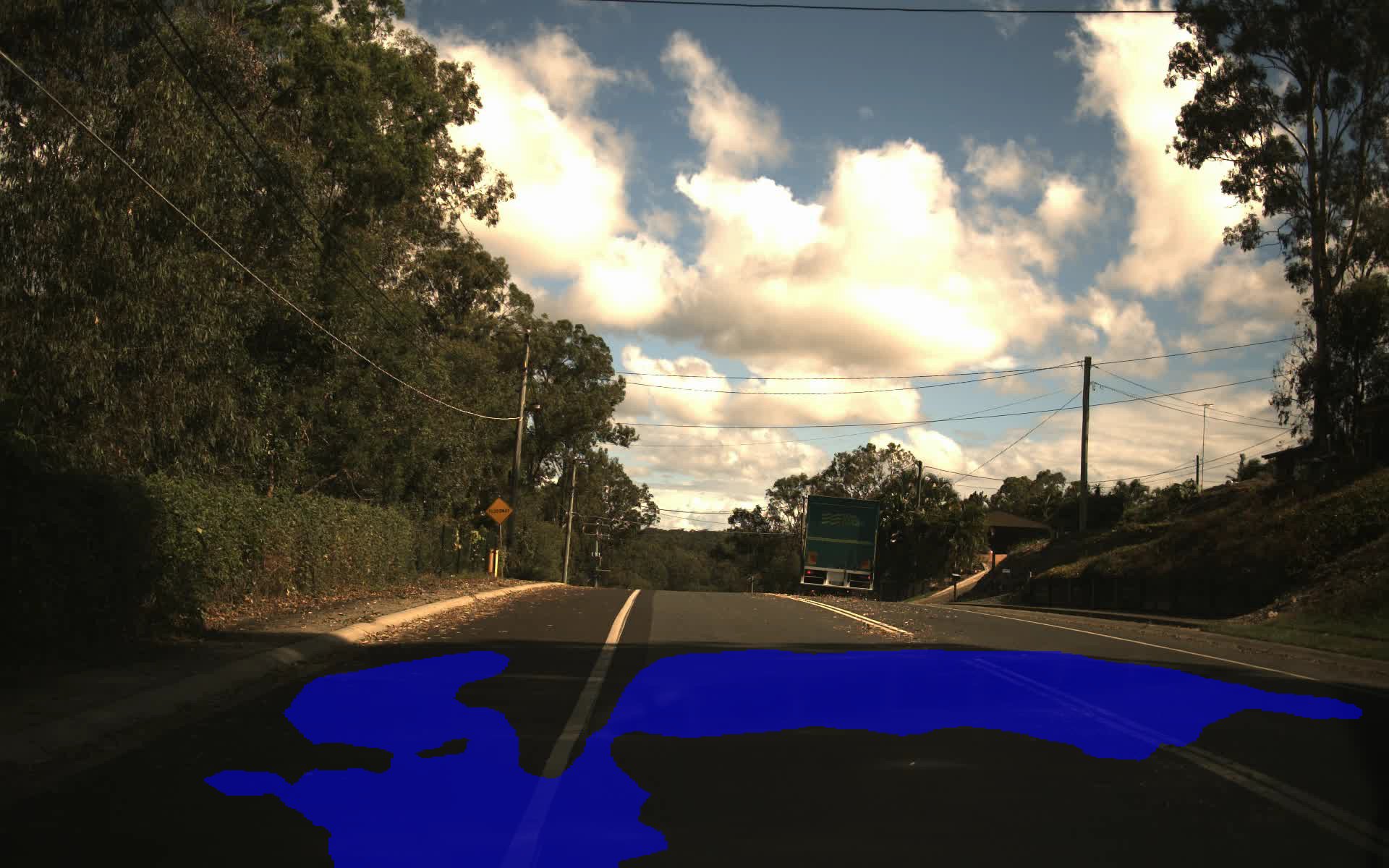} &
        \includegraphics[width=0.49\linewidth]{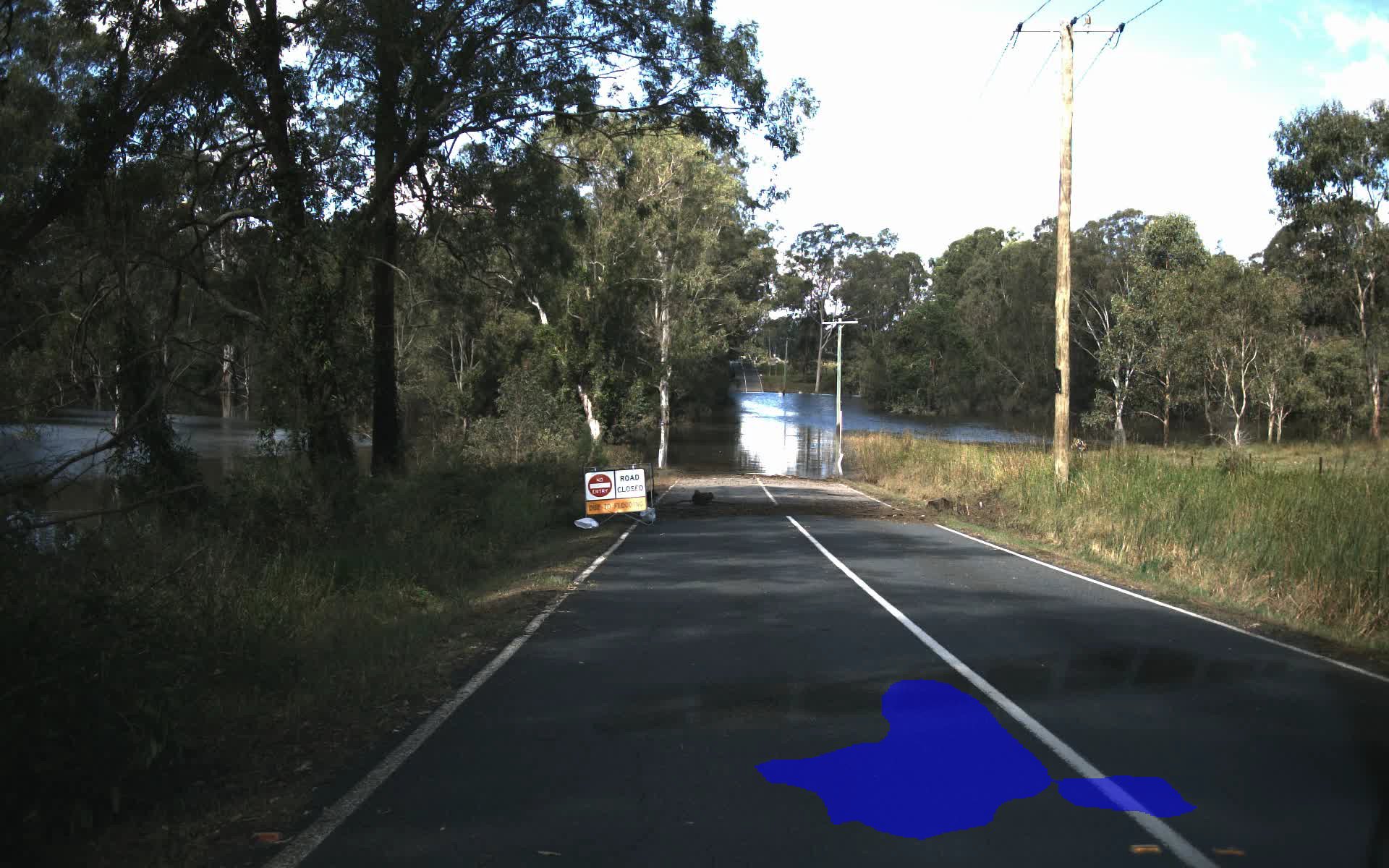}  
    \end{tabular}
    \caption{False positive detections of water hazards on the road can also be problematic during autonomous vehicle operation resulting in phantom braking. Shadows and non-uniform road surfaces were found cause false positives in existing methods. The examples above show Deeplab V3 RAU performance on the Holmview (\textbf{Left}) and Dairy Creek (\textbf{Right}) sequences.}
    \label{fig:falsePositives}
\end{figure}

The other significant challenge identified during evaluation was the frequency of False Positive detections when no water hazards were present in an image. Across the various sequences in the FRED dataset, it was observed that False Positives often occurred on non-uniform road surfaces or when shadows altered the road surface appearance (Figure~\ref{fig:falsePositives}). This can be a critical issue in autonomous vehicle operation that results in phantom braking episodes. Accordingly, it is important that future research considers False Positives during evaluation.

The FRED dataset provides a range of scenarios and conditions that will enable researchers to address these challenges and develop methods that are more robust during deployment.

\subsection{Visual Place Recognition}
\subsubsection{Overview}\ \\
Visual Place Recognition (VPR) is the task of localizing a platform solely from visual data (images). It is generally formulated as an image retrieval problem, where a query image of the current location is compared to a reference database of geo-tagged images, to determine the platform's position within a known environment (\cite{schubert2023visual}). VPR is a fundamental part of autonomous navigation pipelines in vehicles and robots, supporting visual localization/6-DOF pose estimation, as well as loop closure in SLAM systems (\cite{masone2021survey}).

Recently, the success of modern deep-learned descriptor methods has driven more VPR research to investigate their performance outside normal/ideal\footnote{well-illuminated, minimal artefacts due to weather, etc.} conditions (\cite{molloy2020intelligent, waheed2022switchhit, lu2024cricavpr, Malone_2025_ICCV}). However, none of this existing work explores the performance of VPR methods in the presence of large water hazards such as flooded roads. Therefore, in this section, we evaluate a large selection of the most recent and state-of-the-art VPR descriptors on the proposed FRED dataset.

The VPR task is typically defined using the following formulation. Given the descriptor for a query image $q \in \mathbb{R}^D$, and a set of descriptors from the reference database $R = \{ r_i \in \mathbb{R}^D \}_{i=1}^N$, the goal of VPR is to identify the reference image whose descriptor is most similar to the query according to a given distance function $d()$:

\begin{equation}
    \hat{i} = \arg\min_{i \in \{1,\dots,N\}} d(q, r_i)
\end{equation}

The predicted location of the query image is then given by the reference image $r_{\hat{i}}$. In this formulation, $D$ is the dimensionality of the descriptor, $N$ is the number of images in the reference database, and the distance function is typically chosen to be either Euclidean or Cosine distance.

\subsubsection{Metrics: Recall@1}\ \\
The recall@1 metric is commonly used to evaluate place recognition performance. In VPR, the recall@1 is considered identical to the precision at 100\% recall (\cite{schubert2023visual}). That is, it is effectively the percentage of queries where the most similar reference image, with respect to the mathematical distance between descriptors, is considered the same place as the query. Accordingly, a higher recall@1 value indicates higher performance. With the assumption that every query has a corresponding reference image, the recall@1 is calculated by:

\begin{equation}
    Recall@1=\frac{TP}{TP+FP} \ \ \ .
\end{equation}

Where TP (True Positives) is the number of queries matched to the \textit{correct} reference image, and FP (False Positives) is the number of images matched to an \textit{incorrect} reference image.

\subsubsection{Experimental Setup}\ \\
To establish how flooded roads affect VPR performance, we evaluate various VPR descriptors on both the `dry' and `flooded' conditions at all locations in the FRED dataset except Mount Cotton\footnote{Only a `flooded' sequence was recorded for Mount Cotton so there is no sequence to use a reference database.}. This included the following VPR descriptors, BoQ (\cite{ali2024boq}), Clique-Mining (\cite{izquierdo2024close}), CosPlace (\cite{berton2022rethinking}), CricaVPR (\cite{lu2024cricavpr}), EigenPlaces (\cite{berton2023eigenplaces}), MixVPR (\cite{ali2023mixvpr}), SALAD (\cite{izquierdo2024optimal}), and SuperVLAD (\cite{lu2024supervlad}).

A single reference database is created by combining images from one of the `dry' sequences from each respective location. Reference images are taken from sequences:
\begin{itemize}
    \item[] `Cambogan\_20250812\_122339'
    \item[] `Dairy-Creek\_20250812\_122954'
    \item[] `Holmview\_20250812\_120100'
    \item[] `Pullenvale\_20250812\_134316'
\end{itemize}

Flooded condition query sequences include:
\begin{itemize}
    \item[] `Cambogan\_20250811\_113017'
    \item[] `DairyCreek\_20250811\_103318'
    \item[] `Holmview\_20250820\_130327'
    \item[] `Pullenvale\_20250916\_124105'
\end{itemize}

and dry condition query sequences include:
\begin{itemize}
    \item[] `Cambogan\_20250812\_122101'
    \item[] `Dairy-Creek\_20250812\_123312'
    \item[] `Holmview\_20250812\_120856'
    \item[] `Pullenvale\_20250812\_134524'
\end{itemize}

VPR literature is not strictly consistent with the distance tolerance for a reference image to be considered the same place as a query. Some works use a tolerance of $25m$, whereas others use a much tighter tolerance of $\approx 1m$. The purpose of this evaluation is to determine the relative effect of water hazards on VPR performance, not to determine the highest performing VPR descriptor. Therefore, a moderate distance tolerance of $10m$ is used for evaluation. Any query without a corresponding reference image within this distance tolerance was not included in the recall@1 calculation. For the mathematical distance between descriptors, Cosine distance was used.

\begin{table*}
  \centering
  \caption{Recall@1 results for a range of state-of-the-art VPR descriptors on the FRED dataset.}
      \renewcommand\arraystretch{0.89}
      \renewcommand\tabcolsep{3pt}
      \scriptsize
      \begin{tabular}{llcccccccc}
        \toprule[0.03cm]
        \multirow{2}{*}{\textbf{Query Dataset}}
        & \multirow{2}{*}{\textbf{Condition}}
        & \multicolumn{8}{c}{\textbf{VPR Descriptor}} \\
        \cmidrule(lr){3-10}
        & & BoQ & Clique Mining & CosPlace
          & CricaVPR & EigenPlaces & MixVPR & SALAD & SuperVLAD \\
        \midrule[0.03cm]

        Cambogan & Dry & 100 & 100 & 100 & 100 & 100 & 100 & 100 & 100 \\

        Cambogan & Flooded & 87.24 & 85.19 & 83.95 & 61.32 & 87.24 & 86.01 & 90.12 & 84.77 \\

        \cmidrule(lr){1-10}

        Dairy Creek & Dry & 100 & 100 & 100 & 100 & 100 & 100 & 100 & 100 \\

        Dairy Creek & Flooded & 94.29 & 93.57 & 93.21 & 84.29 & 86.43 & 90.00 & 86.79 & 83.93 \\
        
        \cmidrule(lr){1-10}
        
        Holmview & Dry & 100 & 100 & 100 & 100 & 100 & 100 & 100 & 100 \\

        Holmview & Flooded & 99.17 & 99.45 & 99.72 & 81.49 & 98.62 & 98.07 & 98.07 & 99.17 \\
        
        \cmidrule(lr){1-10}
        
        Pullenvale & Dry & 100 & 100 & 100 & 100 & 100 & 100 & 100 & 100 \\

        Pullenvale & Flooded & 100 & 99.67 & 96.66 & 100 & 99 & 99.33 & 99.33 & 100 \\

        \midrule[0.03cm]
        \multicolumn{2}{l}{\textbf{Avg. Reduction (Dry -- Flooded)}} & 4.83 & 5.53 & 6.62 & 18.23 & 7.18 & 6.65 & 6.42 & 8.03 \\

        \bottomrule[0.03cm]
      \end{tabular}
    \label{tab:vpr_results}
    \vspace{-0.2cm}
\end{table*}

\subsubsection{Results and Discussion}\ \\
Table~\ref{tab:vpr_results} demonstrates the reduction in recall@1 performance experienced by each of the VPR descriptors when faced with a flooded road. It can be seen that every VPR descriptor is able to achieve $100\%$ recall@1 across all of the `dry' condition query sequences. This is likely a result of sequences being relatively short (30 seconds) and both the query and reference `dry' sequences being recorded on the same days.

\begin{figure}[h]
    \centering
    \renewcommand\tabcolsep{3pt}
    \begin{tabular}{cc}
        \includegraphics[width=0.49\linewidth]{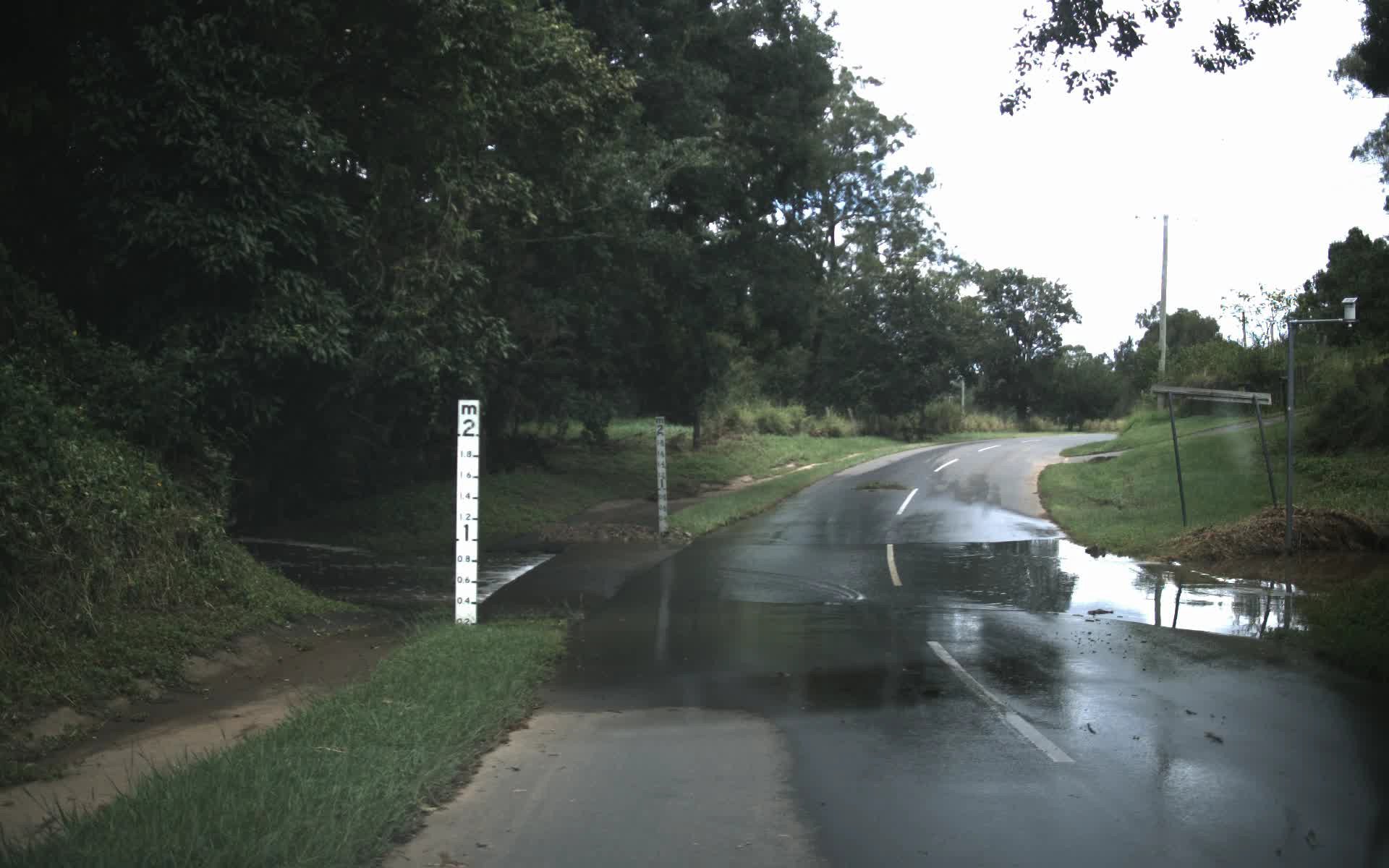} &
        \includegraphics[width=0.49\linewidth]{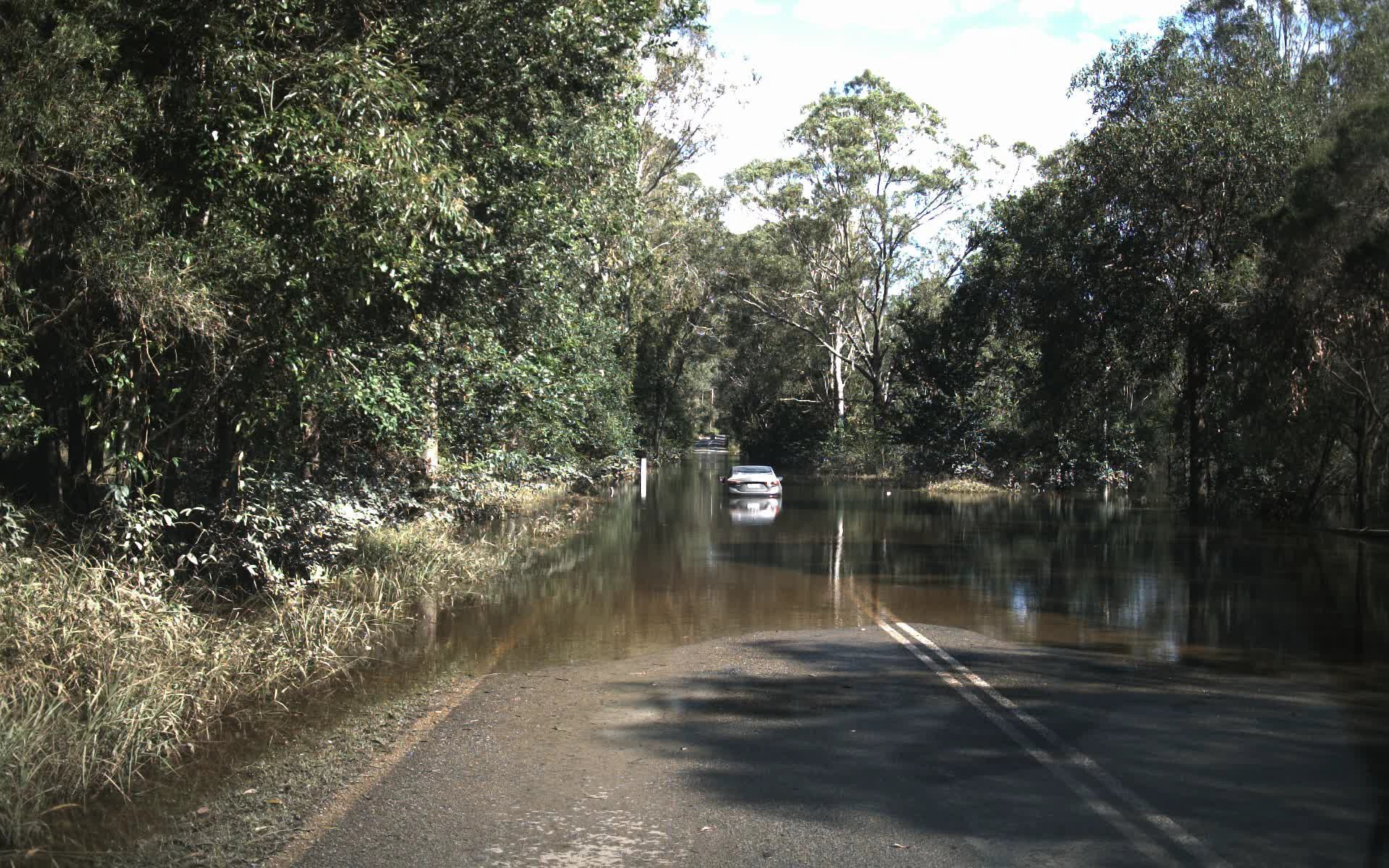}  
    \end{tabular}
    \caption{VPR performance is least affected by the conditions captured in the Pullenvale sequence. This is likely due to the relatively small changes to the images compared to other sequences. \textbf{Left:} Pullenvale. \textbf{Right:} Cambogan.}
    \label{fig:pullcam}
    \vspace{-\baselineskip}
\end{figure}

The table shows that VPR descriptors all typically experience a $5\%$ to $8\%$ reduction in recall@1 on average across the `flooded' query sequences. CricaVPR was particularly affected by the flooded road conditions and experienced an average reduction in recall@1 of $\approx 18\%$. However, the effect on VPR performance was not uniform across all query sequences. For example, all descriptors maintained high VPR performance across the Pullenvale sequence under flooded conditions, whereas, the Cambogan sequence significantly decreased recall@1. This is likely caused by a reduced quantity of water flooding the road in the Pullenvale sequence compared to Cambogan (Figure~\ref{fig:pullcam}). Ultimately, this supports the need for more flooded roads datasets to enable the development of visual localization methods that are robust across different water hazards.

\section{Conclusion}
\label{sec:conclusion}
We have presented, to our knowledge, the first multi-modal autonomous driving dataset focusing on scenarios including water hazards. The Flooded Road Environments Dataset (FRED) includes driving sequences captured from five separate locations in both dry and flooded conditions. Each sequence contains images from a front and rear camera, a 360\textdegree \ LiDAR, and position information from a GNSS corrected IMU. To encourage further research into perception and localization tasks in these scenarios, we provide a development kit with tools for using and visualising the data. Through evaluation of state-of-the-art image-based segmentation and visual place recognition methods, we were able to establish that water hazards present a significant challenge to current methods in both perception and localization. Given the significant lack of publicly available datasets focusing on flooding and water hazards, we hope the release of the FRED dataset enables more research and development for this task.

\begin{acks}
    We would like to thank Amit Trivedi (Queensland Department of Transport and Main Roads), Fatemeh Ghorbani (Queensland University of Technology), and Long Wang (Queensland University of Technology) for their contribution to this work through iMove project \mbox{1-075} (\textit{Expanding Operating Design Domain of Automated Vehicles}).
\end{acks}

\begin{dci}
    The author(s) declared no potential conflicts of interest with respect to the research, authorship, and/or publication of this article.
\end{dci}

\begin{funding}
The author(s) disclosed receipt of the following financial support for the research and publication of this article: This work was supported by the The Queensland Department of Transport and Main Roads, The iMove Cooperative Research Centre (CRC) through the project, \textit{Expanding Operating Design Domain of Automated Vehicles}, (project 1-075), and The Australian Research Council (ARC) through the ARC Industrial Transformation Training Centre for Automated Vehicles in Rural and Remote Regions (IC230100001).
\end{funding}

\bibliographystyle{SageH}
\bibliography{references}

\end{document}